%% file: iclr2025_conference.tex
\documentclass{article} 
\usepackage{iclr2025_conference,times}
\iclrfinalcopy

\input{math_commands.tex}

\usepackage{hyperref}
\usepackage{url}
\usepackage{graphicx}
\usepackage{booktabs}       
\usepackage{amsfonts}       
\usepackage{nicefrac}       
\usepackage{microtype}      
\usepackage{xcolor}         

\usepackage{amsmath,amssymb} 
\usepackage{color}
\usepackage[noend]{algpseudocode}
\usepackage[linesnumbered,vlined,ruled,commentsnumbered]{algorithm2e}

\usepackage[accsupp]{axessibility}  

\usepackage{adjustbox}
\usepackage{colortbl}

\usepackage{multirow, booktabs, tabularx}
\usepackage{subcaption}
\usepackage{caption}

\usepackage{wrapfig}

\SetKwInput{KwInput}{Input}       
\SetKwInput{KwOutput}{Output}     

\title{Prompting the Unseen: Detecting Hidden Backdoors in Black-Box Models}

\author{Zi-Xuan, Huang, Jia-Wei Chen, Zhi-Peng Zhang, Chia-Mu Yu \\
National Yang Ming Chiao Tung Univeraity\\
Hsinchu City 300093, Taiwan \\
\texttt{\{reborn.1998.wong,jiawei,changchihpeng911,chiamuyu\}@gmail.com} \\
}

\begin{document}

\maketitle

\input{sec/0_abstract}    
\input{sec/1_intro}
\input{sec/2_related}

\input{sec/3_background}

\input{sec/4_proposed}
\input{sec/5_experiment}

\input{sec/6_conclusion}


\bibliography{iclr2025_conference}
\bibliographystyle{iclr2025_conference}

\clearpage
\appendix
\input{sec/x_appendix}

\end{document}

%% file: math_commands.tex
\usepackage{amsmath,amsfonts,bm}

\def\eqref#1{equation~\ref{#1}}

\def\1{\bm{1}}

\DeclareMathAlphabet{\mathsfit}{\encodingdefault}{\sfdefault}{m}{sl}
\SetMathAlphabet{\mathsfit}{bold}{\encodingdefault}{\sfdefault}{bx}{n}



%% file: sec/0_abstract.tex
\begin{abstract}
Visual prompting (VP) is a new technique that adapts well-trained frozen models for source domain tasks to target domain tasks. This study examines VP's benefits for black-box model-level backdoor detection. The visual prompt in VP maps class subspaces between source and target domains. We identify a misalignment, termed class subspace inconsistency, between clean and poisoned datasets. Based on this, we introduce \textsc{BProm}, a black-box model-level detection method to identify backdoors in suspicious models, if any. \textsc{BProm} leverages the low classification accuracy of prompted models when backdoors are present. Extensive experiments confirm \textsc{BProm}'s effectiveness.
\end{abstract}

%% file: sec/1_intro.tex
\section{Introduction}
\label{sec: Introduction}
Deep neural networks (DNNs) are commonly used in complex applications but require extensive computational power, leading to significant costs. Users often access these models through online platforms like BigML model market\footnote{\url{https://bigml.com/}} and ONNX zoo\footnote{\url{https://github.com/shaoxiaohu/model-zoo}}, or via Machine Learning as a Service (MLaaS) platforms. However, DNNs can include backdoors~\citep{Gu2017BadNetsIV,Liu2018TrojaningAO,scan, revisitingbackdoor,Nguyen2021WaNetI,Chen2017TargetedBA}, which manipulate model responses to inputs with specific triggers (like certain pixel patterns) while functioning correctly on other inputs. In backdoor attacks, attackers embed these triggers in the training data, leading the model to associate the trigger with a particular outcome and misclassify inputs containing it.
\vspace{-0.3cm}
\paragraph{Why Black-Box Model-Level Detection.}\label{sec: Why Black-Box Backdoor Detection.}
Black-box backdoor detection, which uses only black-box queries to the suspicious model (i.e., the model to be inspected), is gaining attention. This detection method is divided into input-level~\citep{iclrworkshop2022, deepsweep, 9343758, teco, ct, metasift, scaleup,pmlr-v235-hou24a,xu2024towards,ted} and model-level~\citep{onepixel, Dong_2021_ICCV, aeva, mntd,mmbd} techniques. Input-level detection identifies trigger samples in an infected model, while model-level detection determines if a model contains backdoors. Input-level detection relies on the model having backdoors; otherwise, its accuracy drops significantly. For example, as shown in Table~\ref{tbl:drop}, TeCo~\citep{teco} and SCALE-UP~\citep{scaleup}, state-of-the-art input-level detectors, show AUROCs of 0.8113 and 0.7877, respectively, on a BadNets-infected model~\citep{Gu2017BadNetsIV}, but only 0.4509 and 0.5103 on a clean model. If a model is clean, many legitimate samples may be misclassified as triggers, reducing the model's practical utility. Thus, model-level detection should be performed first. If backdoors are found but the model must still be used, input-level detection should then be applied to each input.

\vspace{-0.2cm}
\begin{table}[htbp]
\centering
\caption{A significant drop of F1-score and AUROC in black-box input-level detection methods, TeCo~\citep{teco} and SCALE-UP~\citep{scaleup}.}
\vspace{0.1cm}

\begin{adjustbox}{max width=0.85\columnwidth}
\begin{tabular}{@{} lcccccc @{}}
\toprule
& \multicolumn{2}{c}{BadNet~\citep{Gu2017BadNetsIV}} & \multicolumn{2}{c}{Blended~\citep{Chen2017TargetedBA}} & \multicolumn{2}{c}{WaNet~\citep{Nguyen2021WaNetI}} \\
\cmidrule(r){2-3} \cmidrule(l){4-5} \cmidrule(l){6-7}
TeCo~\citep{teco} & Backdoored & Clean & Backdoored & Clean & Backdoored & Clean \\
\midrule
F1    & 0.8014 & 0.5263 & 0.7621 & 0.5033 & 0.9295 & 0.5137 \\
AUROC & 0.8113 & 0.4509 & 0.7259 & 0.3954 & 0.9345 & 0.4406 \\
\addlinespace
\hline
\hline
\addlinespace
\addlinespace
ScaleUp~\citep{scaleup} & Backdoored & Clean & Backdoored & Clean & Backdoored & Clean \\
\midrule
F1    & 0.7964 & 0.5236 & 0.7991 & 0.5046 & 0.7199 & 0.4768 \\
AUROC & 0.7877 & 0.5103 & 0.7694 & 0.4643 & 0.7772 & 0.4246 \\
\addlinespace
\hline
\hline
\end{tabular}
\end{adjustbox}
\vspace{-0.3cm}
\label{tbl:drop}
\end{table}

\vspace{-0.2cm}
\paragraph{Design Challenge.}\label{sec: Design Challenge.} Despite its importance, black-box model-level detection faces two main challenges. First, unlike input-level detection, which benefits from the presence of an infected model, model-level detection has limited ground truth, relying on only a few clean samples. Second, it needs a stable feature to differentiate between clean and infected models across various backdoor types, which is difficult to find. For instance, B3D~\citep{Dong_2021_ICCV} targets trigger localization but is mainly effective for patch-based triggers. Similarly, AEVA~\citep{aeva} may struggle with larger triggers due to its dependence on adversarial peak analysis.

\begin{figure}
     \centering
     \begin{subfigure}[t]{0.47\textwidth}
         \centering
         \includegraphics[width=\textwidth]{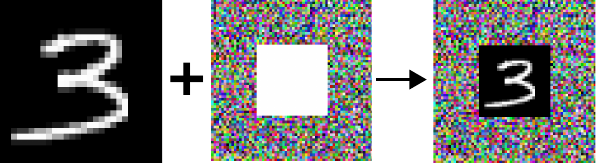}
         \caption{The ``3'' is from MNIST, while the middle part shows the visual prompt. The prompted sample is ready for ImageNet classifier.}
         \label{fig: 1a}
     \end{subfigure}
     \hfill
     \begin{subfigure}[t]{0.47\textwidth}
         \centering
         \includegraphics[width=\textwidth]{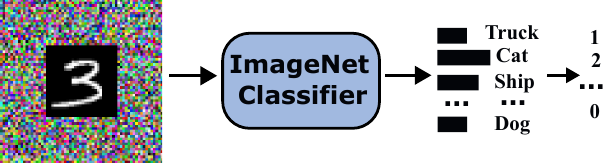}
         \caption{The prompted sample can be fed into the ImageNet classifier, whose output has a mapping between the labels from MNIST and ImageNet.}
         \label{fig: 1b}
     \end{subfigure}
\caption{How a frozen ImageNet classifier is adapted for the MNIST classification when VP is used.}
\vspace*{-.3cm}
\label{fig: 1}
\end{figure}

\vspace{-0.3cm}\paragraph{Our Design.} Visual prompting (VP)~\citep{vporiginal,vpt} allows a frozen, pre-trained model from a source domain to correctly predict samples from a target domain by applying a visual prompt. This technique can work across very different domains; for example, an ImageNet classifier (source) can detect melanoma (target) via VP~\citep{yunyuntsai}. Figure~\ref{fig: 1} illustrates VP, where the visual prompt (trainable noise in Figure~\ref{fig: 1a}) maps between class subspaces of the source and target domains, enabling the frozen classifier to handle the target task efficiently.

\begin{figure*}
      \centering
      \begin{subfigure}[b]{0.47\textwidth}
          \centering
          \includegraphics[width=\textwidth]{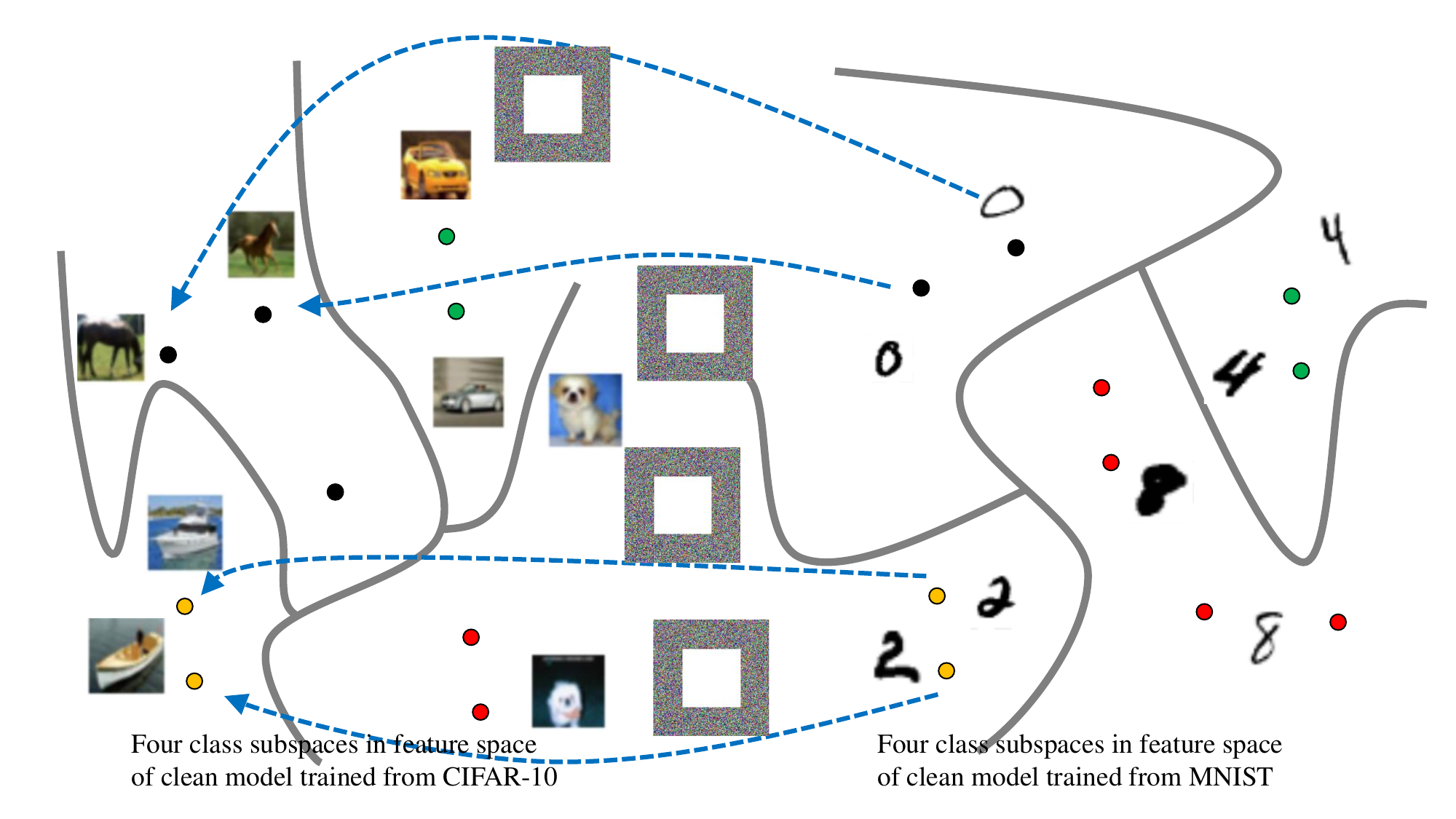}
          \caption{Class subspace inconsistency does not occur: visual prompt as a mapping between two clean datasets.}
          \label{fig:2a}
      \end{subfigure}
      \hfill
      \begin{subfigure}[b]{0.47\textwidth}
          \centering
          \includegraphics[width=\textwidth]{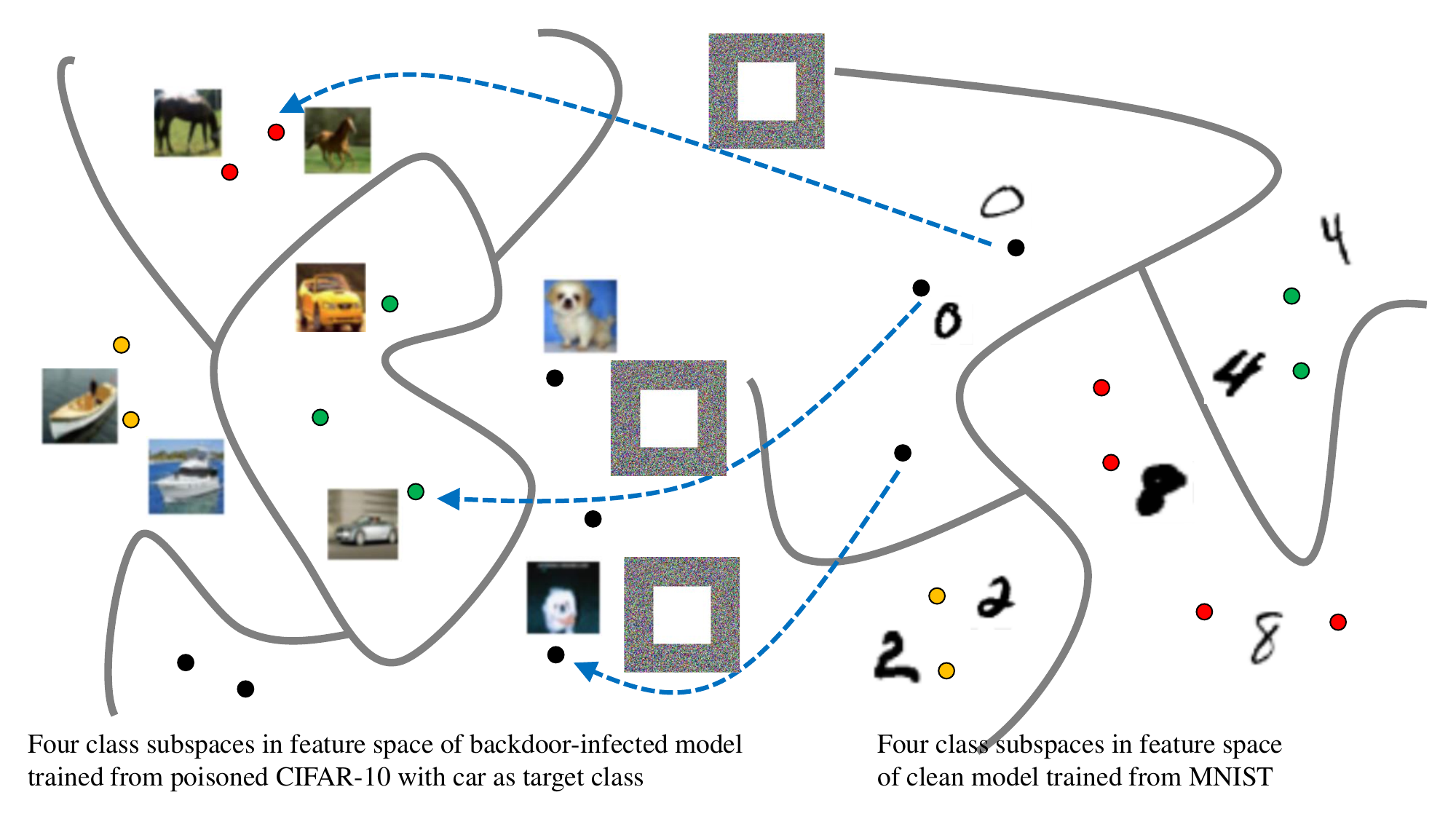}
          \caption{Subspace inconsistency occurs: visual prompt as a mapping between clean and poisoned datasets.}
          \label{fig:2b}
      \end{subfigure}
 \caption{A conceptual illustration of (a) VP on clean model and (b) VP on backdoor-infected model.}
 \vspace*{-.3cm}
 \label{fig: class subspace inconsistency}
 \end{figure*}

\begin{figure}[t]
    \centering
    \begin{subfigure}[t]{0.43\textwidth}
        \includegraphics[width=\textwidth]{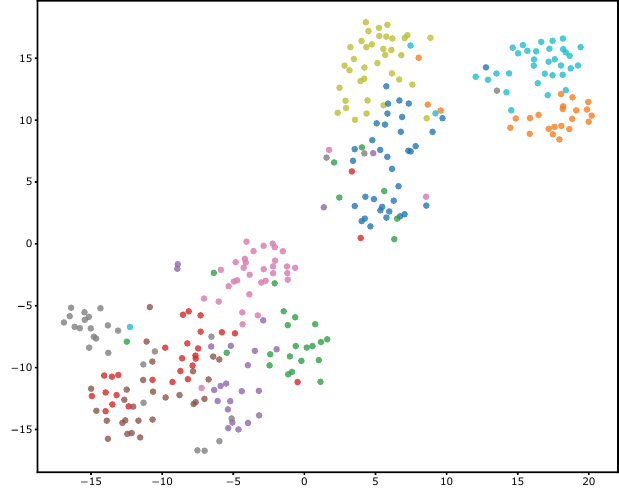}
        \caption{\scriptsize Clean source model shows clear class separation.}
        \label{fig:clean_source}
    \end{subfigure}
    \begin{subfigure}[t]{0.43\textwidth}
        \includegraphics[width=\textwidth]{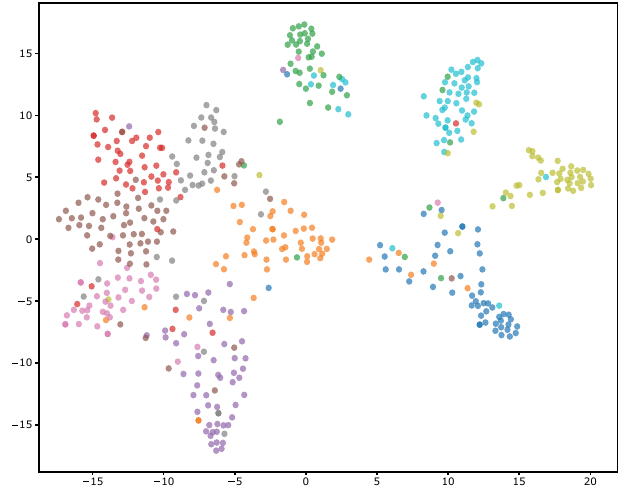}
        \caption{\scriptsize Clean target model preserves clear separation.}
        \label{fig:clean_target}
    \end{subfigure}

    \begin{subfigure}[t]{0.43\textwidth}
        \includegraphics[width=\textwidth]{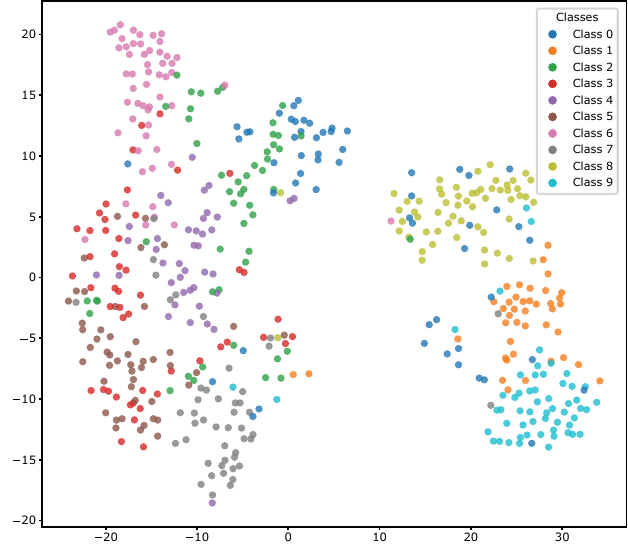}
        \caption{\scriptsize Target class (0) is adj. to others in infected source model.}
        \label{fig:3c}
    \end{subfigure}
    \begin{subfigure}[t]{0.43\textwidth}
        \includegraphics[width=\textwidth]{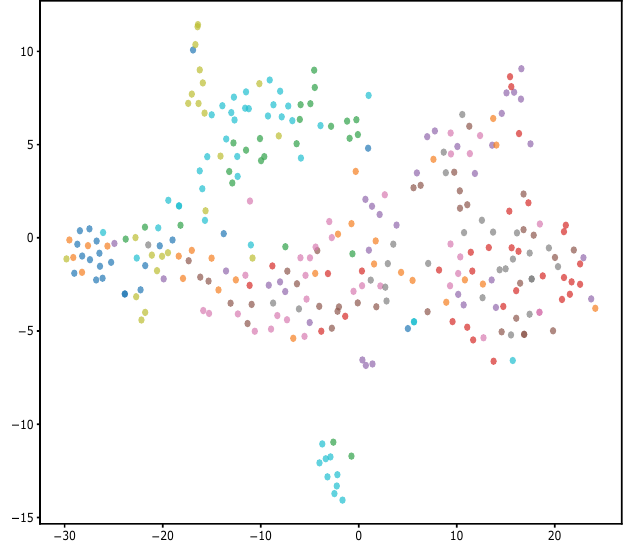}
        \caption{\scriptsize Infected target model shows severe class confusion.}
        \label{fig:infected_target}
    \end{subfigure}
    \caption{Class subspaces inconsistency (CIFAR-10 for source model and STL-10 for target model). }
    \vspace*{-0.5cm}
    \label{fig:class_subspace_visualization}
\end{figure}

In an infected model, the target class subspace in the feature space is adjacent to all other class subspaces~\citep{Wang2019NeuralCI}. We identify a \textit{class subspace inconsistency} where misalignment between class subspaces in the poisoned (source) and clean (target) datasets leads to low classification accuracy of the prompted model. This phenomenon is illustrated in Figure~\ref{fig: class subspace inconsistency} and experimentally validated in both Figure~\ref{fig:class_subspace_visualization} and Section~\ref{sec: Another Visualization of Class Subspace Inconsistency}. Based on this, we propose \textsc{BProm} for black-box model-level backdoor detection. \textsc{BProm} applies VP to a suspicious model using an unrelated clean dataset; poor accuracy in the prompted model indicates the presence of backdoors. 

\textbf{Contribution.} Our contributions can be summarized as follows. 1) We identify a \textit{class subspace inconsistency} in VP on backdoor-infected models. This misalignment between class subspaces of the poisoned dataset and an external clean dataset signals backdoor infection. 2) Utilizing this inconsistency, we develop \textsc{BProm}, a black-box model-level backdoor detection method.

%% file: sec/2_related.tex
\section{Related works}\label{sec: Related works}
\label{sec:related}
We do not aim to provide a comprehensive review of backdoor attacks and defenses; for a detailed survey, see \citep{9802938}.

\paragraph{Backdoor Attack Methods.}
Badnets~\citep{Gu2017BadNetsIV} introduced the first backdoor attack on DNNs, with many following works adopting its approach to poison training datasets. Backdoor attacks are categorized by trigger appearance into universal~\citep{Chen2017TargetedBA, Gu2017BadNetsIV, low-frequency}, where all triggers are identical, and sample-specific~\citep{Li2020InvisibleBA, input-aware, dynamic}, where triggers vary per sample. Subsequent developments include invisible backdoors~\citep{lira, Li2020InvisibleBA, Nguyen2021WaNetI}, which are harder to detect by human inspection, and clean-label backdoors~\citep{Zhao_2020_CVPR, 9488902, poisonfrog, cleanlabel}, which stealthily poison target class samples without label changes. Additionally, anti-defense attacks~\citep{revisitingbackdoor} circumvent detection by preventing latent separation.
\paragraph{Backdoor Detection Methods.}
Backdoor detection is categorized into white-box and black-box. White-box detection~\citep{finepruning, anp, Li2021NeuralAD, ijcai2022p206, Du2020Robust, li2021anti, huang2022backdoor, Wang2019NeuralCI, hu2022trigger, 9879000, wei2024shared, li2023reconstructive, mmbd} requires access to a poisoned training set or model parameters. Some methods identify backdoors, while others remove them. However, it is unsuitable for MLaaS applications and safety-critical deployments (e.g., autonomous vehicles).

Black-box detection only requires access to the suspicious model, making it more applicable. It is divided into input-level and model-level. Input-level detection~\citep{iclrworkshop2022, deepsweep, 9343758, teco, ct, metasift, scaleup, xian2024unified, ma2022beatrix, pan2023asset, jin2022incompatibility, chen2024progressive, zhu2024neural,pmlr-v235-hou24a,xu2024towards} distinguishes trigger samples from benign ones. Since infected models act benign except for trigger samples, they can be used safely if detection works per input. However, this can result in high false positives, rejecting many benign samples if the model is clean, as shown in Table~\ref{tbl:drop}.

This paper focuses on model-level detection~\citep{onepixel, Dong_2021_ICCV, aeva, mntd, shi2024black, xiang2024cbd, sunneural, rezaei2023run,mmbd}, which identifies backdoors in suspicious models and serves as front-line detection before input-level methods.

%% file: sec/3_background.tex
\section{Background Knowledge}\label{sec: Background Knowledge}

Both visual prompting (VP)~\citep{Chen_2023_CVPR, vporiginal,vpt} and model reprogramming (MR)~\citep{yunyuntsai, Chen2022ModelRR, Elsayed2018AdversarialRO, Neekhara_2022_WACV} enable a frozen pre-trained model for one task to perform a different target domain classification task by deriving a visual prompt for inputs from the target domain. Initially, MR was considered an \textit{attack} that misused cloud services (i.e., MLaaS) to perform undocumented tasks~\citep{Elsayed2018AdversarialRO}. VP was recently introduced in \citep{vporiginal}. Although VP and MR share the same concept, VP focuses exclusively on images. VP has been extended to image inpainting~\citep{vpinpainting}, antibody sequence infilling~\citep{reprogrammingforbio}, and differentially private classifiers~\citep{prompate}. In this paper, VP and MR are used interchangeably, with the visual prompt in VP corresponding to the trainable noise in MR. More formally, VP/MR proceeds with four steps \citep{Chen2022ModelRR}. 

1. \underline{Initialization}: Let $f_S(\cdot)$ and $D_T=\{ (x_T, y_T)\}$ be the source model (the model trained from the source domain dataset) and the target domain dataset, respectively. Randomly initialize $\theta$ and $w$ (defined below).

2. \underline{Visual prompt padding}: Obtain the prompted input sample $\tilde{x}_T = V(x_T|\theta)$, where $\theta$ is the visual prompt. A common method for $V(\cdot)$ is to resize $x_T$ and add the visual prompt (trainable noise) around it. Although $\tilde{x}_T$ visually differs from the source domain, it can still be used as input for the source domain classifier. Figure~\ref{fig: 1a} illustrates this with $x_T$ as ``3'' from MNIST, $\theta$ in the middle, and $V(\cdot)$ resizing $x_T$ and padding it with $\theta$.

3. \underline{Output mapping}: Obtain the target task prediction via $\hat{y}_T = O(f_S(\tilde{x}_T) | w)$, where $w$ represents the trainable parameters for output label mapping. This step is optional for VP/MR. In our experiment, we omitted this step.

4. \underline{Prompted model training}: Optimize $\theta$ and $w$ by minimizing a task-specific loss $\mathcal{L}(\hat{y}_T, y_T)$ on $D_T$. 

After executing the four-step procedure, we obtain the prompted model $f_T=O\circ f_S\circ V$ from $f_S(\cdot)$ with optimized $\theta^*$ and optionally $w^*$. This results in $\hat{y}_T = O(f_S(V(x_T|\theta^*))|w^*)$.

\section{System Model}\label{sec: System Model}
\paragraph{Threat Model.}\label{sec: Threat Model.}
We consider two roles: attacker and defender. The attacker's goal aligns with previous work~\citep{Gu2017BadNetsIV, Chen2017TargetedBA, scan, revisitingbackdoor,Liu2018TrojaningAO}. Specifically, the attacker poisons the training dataset by injecting trigger samples. The DNN model (e.g., an image classifier) trained on this poisoned dataset behaves normally with clean inputs but always predicts an attacker-specified target class for inputs with a trigger. Essentially, an all-to-one backdoor is implanted, mapping all trigger inputs to a specific target class.

\paragraph{Defender's Goal and Capability.}\label{sec: Defender's Goal and Capability}
The defender's goal is to detect if a suspicious model is backdoored, primarily measured by AUROC (see Section~\ref{sec: Experiments}). The defender has limited abilities: no access to the poisoned dataset, model structure, or parameters. In MLaaS applications, detection involves only black-box queries on the model to obtain confidence vectors. The defender also has a small reserved clean dataset $D_S$ (1\%, 5\%, 10\% of the test dataset in our experiment) to aid detection.

%% file: sec/4_proposed.tex
\begin{figure*}[t]
    \centering
    \includegraphics[width=1.02\linewidth]{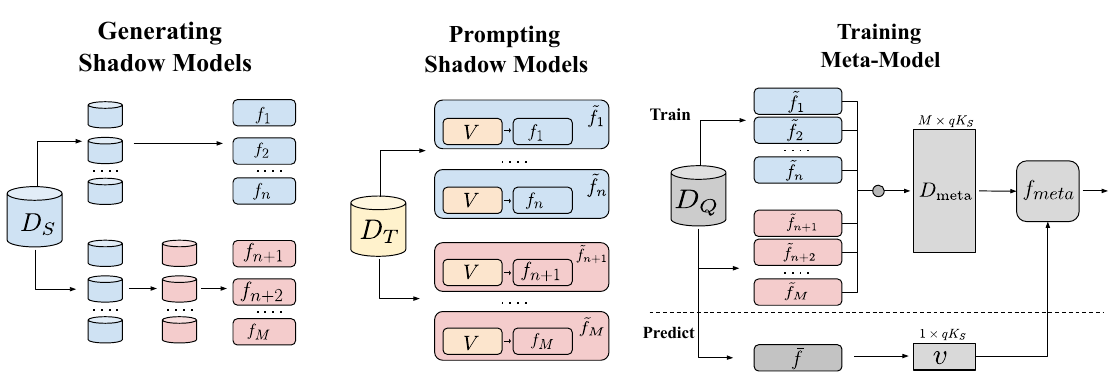}
    
\caption{The workflow of \textsc{BProm}. The \textcolor[rgb]{0.00,0.00,1.00}{blue} and \textcolor[rgb]{1.00,0.00,0}{red} components are related to $D_S$ and $D_P$, respectively. The \textcolor[rgb]{1.00,1.00,0.50}{yellow} parts are related to VP and $D_T$. The \textcolor[rgb]{0.35,0.43,0.46}{gray} components have connection to $D_Q$ and are used for train $f_{\text{meta}}$}. 
\vspace*{-.3cm}
\label{fig:overview}
\end{figure*}

\section{Proposed Method}\label{sec: Proposed Method}
We present our detection method, \textsc{BProm}. The notation table can be found in Table~\ref{tab:notation} in Appendix~\ref{sec: Notation and Definitions}. 

\subsection{Overview}\label{sec: Overview}


Different clean datasets have distinct class subspace "shapes" in feature space. However, as noted in \citet{Wang2019NeuralCI}, poisoned datasets exhibit target class subspaces that share boundaries with all others. This creates misalignment when adapting a poisoned model to a clean dataset, termed \textit{class subspace inconsistency}, resulting in reduced prompted model accuracy. This is conceptually illustrated in Figure~\ref{fig: class subspace inconsistency} and experimentally validated in Section~\ref{sec: Another Visualization of Class Subspace Inconsistency} and Figure~\ref{fig:class_subspace_visualization}. As an evidence, Table~\ref{table: evidence-of-Black-Prom} also shows that  an increasing number of target classes worsens the inconsistency (i.e., lower accuracy). \textsc{BProm} leverages this for backdoor detection. The core idea is that adapting an infected source model to a clean target task via visual prompting is significantly harder due to the class subspace mismatch.  Theorem 1 in \citet{yang2021voice2series} states that target risk is bounded by source risk and representation alignment loss. For \textsc{BProm}, this alignment loss is amplified by the inconsistency in infected models, leading to poor target task performance. Thus, low prompted accuracy signals potential backdoors. To achieve effective detection, the \textsc{BProm} training has three steps: shadow model generation, prompting, and meta-model training.  First, diverse poisoned and clean shadow models are trained. Second, visual prompts are learned for each shadow model using an external clean dataset. Finally, a meta-classifier is trained on confidence vectors from prompted shadow models to detect backdoors.  The workflow and pseudocode are shown in Figure~\ref{fig:overview} and Algorithm~\ref{algorithm: main algorithm}.

\begin{wraptable}{r}{7cm}
\centering
\footnotesize
\vspace*{-.3cm}
\caption{Class subspace inconsistency worsens (i.e., the prompted model's testing accuracy decreases) as the number of target classes increases.}
\begin{adjustbox}{max width=0.5\columnwidth}

\begin{tabular}{cccc}
\hline
\parbox{1.5cm}{\textbf{\tiny \# target classes}} & \textbf{1} & \textbf{2} & \textbf{3} \\
\hline
CIFAR10 & 0.3286 & 0.2427 & 0.2338 \\
GTSRB & 0.2711 & 0.1988 & 0.1986 \\
\hline
\end{tabular}
\label{table: evidence-of-Black-Prom}
\end{adjustbox}
\vspace*{-0.3cm}
\end{wraptable}


\subsection{\textsc{BProm}}\label{sec: Black-Prom}

\paragraph{Generating Shadow Models.}\label{sec: Generating Shadow Models}

The goal of this step is to construct shadow models, categorized into clean and backdoor shadow models. Clean shadow models are trained on a clean dataset, while backdoor shadow models are trained on a poisoned dataset.

Let $D_S$ be the reserved clean dataset. To check if a suspicious model was trained on CIFAR-10, $D_S$ includes a limited number of CIFAR-10 samples (e.g., 1\%, 5\%, 10\% in our experiment). The defender trains $n$ clean shadow models, $f_i$'s, with different parameter initializations. Given a poisoning rate $p$ and a chosen backdoor attack, the defender creates $M-n$ poisoned datasets by injecting trigger samples according to the chosen attacks, where $M$ is the total number of shadow models. Specifically, each poisoned dataset $D_P$ is constructed as follows:

\textbf{Step 1:} A proportion $p$ of samples $(x, y)$ from the clean dataset $D_S$ are extracted to form $D_E$.

\textbf{Step 2:} The extracted samples are transformed by adding a trigger pattern ($m, t, \alpha, y_t$) to obtain poisoned counterparts $\{(x', y')|x' = (1 - m) \cdot x + m \cdot ((1 - \alpha)t + \alpha x)$, $y' = y_t\}$, where $y_t$, $m$, $t$, $\alpha$, $\cdot$ denote the target class, trigger mask, trigger, intensity, and element-wise product, respectively~\citep{aeva, scaleup}.

\textbf{Step 3:} Construct $D_P=(D_S \setminus D_E) \cup \{(x', y')\}$. By sampling different combinations of backdoor patterns ($m, t, \alpha, y_t$), various $D_P$ can be generated. Backdoor shadow models are trained on $D_P$'s.

\paragraph{Prompting Shadow Models.}
This step applies VP to both types of shadow models (clean and poisoned) to generate prompted shadow models.  Let $D_T = D_T^{\text{train}} \cup D_T^{\text{test}}$ be an external clean dataset, with $D_T^{\text{train}}$ as the training set and $D_T^{\text{test}}$ as the test set.  $D_T$ can have a different distribution than $D_S$. For shadow models, prompts ($\theta_i$) are learned via standard backpropagation on  $D_T^{\text{train}}$.  This process is also applied to the suspicious model $f_{sus}$, but using a gradient-free optimization method (e.g., CMA-ES) since we only have black-box access.  This results in the prompted shadow models $\tilde{f}_i(\cdot) = f_i(V(\cdot|\theta_i^*))$, and prompted suspicious model $\bar{f}(\cdot) = f_{sus}(V(\cdot|\theta_{sus}^*))$.  Detailed steps for VP can be found in Section~\ref{sec: Background Knowledge} (e.g.,~\citep{vporiginal}).

\begin{wrapfigure}{r}{7cm}
\vspace{-0.4cm}

\begin{adjustbox}{max width=0.5\columnwidth}
\hspace{0.5cm}
\begin{algorithm}[H]
\DontPrintSemicolon
\caption{\textsc{BProm}.}
\label{algorithm: main algorithm}
\KwInput{ 
 $D_S$ and $D_T=D_T^\text{train}\cup D_T^\text{test}$;
}
\KwOutput{$f_{\text{meta}}$}

\tcc{\textcolor[rgb]{0.00,0.00,1.00}{Generating Shadow Models}}
\For{$i=1$ \KwTo $M$}{
    Copy $D_S$ into $D^i_S$\;
    \eIf{$i \leq n$}{
        train $f_i$ from $D_S^i$ \; 
    }
    {
        augment $D^i_S$ with triggers \;
        train $f_i$ from the augmented $D_S^i$ \; 
    }
}
\tcc{\textcolor[rgb]{0.00,0.00,1.00}{Prompting Shadow Models}}
\For{$i=1$ \KwTo $M$}{
    learn visual prompt $\theta_i$ on $D_T^{\text{train}}$\;  
    construct $\tilde{f}_i=f_i\circ V(\cdot|\theta)$\;
}
\tcc{\textcolor[rgb]{0.00,0.00,1.00}{Training Meta Model}}
Construct $D_Q=\{x^1_Q,x^2_Q,...,x^q_Q\}$ by randomly sampling $q$ samples from $D_T^\text{test}$\;
Initialize $D_{\text{meta}}$ as an empty set\;
\For{$i=1$ \KwTo $M$}{
    \eIf{$i \leq n$}{
        $v_i \gets (\tilde{f}_i(x^1_Q)||\cdots||\tilde{f}_i(x^q_Q))$\;
        $l_i \gets \textsf{`clean'}$\;
        $D_{\text{meta}}=D_{\text{meta}}\cup \{(v_i, l_i)\}$\;
    }
    {
        $v_i \gets (\tilde{f}_i(x^1_Q)||\cdots||\tilde{f}_i(x^q_Q))$\;
        $l_i \gets \textsf{`backdoor'}$\;
        $D_{\text{meta}}=D_{\text{meta}}\cup \{(v_i, l_i)\}$\; 
    }
}
Train the binary classifier $f_\text{meta}$ using $D_\text{meta}$\;

\Return{$f_{\text{meta}}$}\;

\end{algorithm}
\end{adjustbox}
\vspace{-2.5cm}
\end{wrapfigure}

\paragraph{Meta Model Training.}\label{sec: meta-training}
The goal of this step is to train a binary classifier $f_{\text{meta}}$ for backdoor detection. For each shadow model $\tilde{f}_i$, the defender randomly selects $q$ samples from $D_T^{\text{test}}$ to form $D_Q=\{x_Q^1, \dots, x_Q^q\}$. Each sample from $D_Q$ is fed to $\tilde{f}_i$. The defender creates a dataset $D_{\text{meta}}=D_{\text{meta}} = \{ (\tilde{f}_i(x_Q^1)||\cdots||\tilde{f}_i(x_Q^q), \textsf{clean}) \}_{i=1}^n \cup \{ (\tilde{f}_i(x_Q^1)||\cdots||\tilde{f}_i(x_Q^q), \textsf{backdoor}) \}_{i=n+1}^M$. Here, $\tilde{f}_i(x_Q^j)$ is the confidence vector, and its length, $K_S$, is the number of classes in $D_S$. The defender then trains a binary classifier $f_{\text{meta}}$ using $D_{\text{meta}}$.

\paragraph{Backdoor Detection on Suspicious Model.}
To inspect a suspicious model $f_{sus}$, we first obtain $q$ confidence vectors from the prompted suspicious model $\bar{f}$. These vectors are concatenated and fed to $f_{\text{meta}}$. Specifically, $v = (\bar{f}(x_Q^1)||\cdots||\bar{f}(x_Q^q))$ is computed and input to $f_{\text{meta}}$, which outputs either \textsf{clean} or \textsf{backdoor}.

\subsection{Discussion}
\textsc{BProm} is similar to MNTD~\citep{mntd}, but they have important differences.

\textbf{More Efficient Data Generation:} In \textsc{BProm}, the defender uses a single backdoor attack to generate $D_P$, whereas MNTD uses multiple backdoor attacks. Even if multiple methods are used in \textsc{BProm}, detection accuracy improves only marginally. MNTD needs to "see" various backdoor types to better detect unknown backdoors. However, \textsc{BProm} focuses on class subspace inconsistency, where $D_P$ learns different feature space partitions, with the target class adjacent to all other classes.

\textbf{Much Fewer Shadow Models Required:} \textsc{BProm} needs only a few shadow models (e.g., 20 in our experiments), while MNTD requires hundreds due to the variety of backdoor attacks (e.g., 256 in MNTD). This reduces training costs and allows \textsc{BProm} to achieve high performance (1.0 AUROC on CIFAR-10 for both BadNets and Blend, compared to MNTD's 0.92 and 0.955) even with a single backdoor type. Training MNTD is also much more complex than training \textsc{BProm}.


\textbf{Novel Design Principle:} Most importantly, their design principles differ fundamentally. MNTD relies on meta-learning and needs to "see" various backdoor properties. \textsc{BProm} relies on class subspace inconsistency, achieving decent detection accuracy (e.g., 0.8137 F1-score on CIFAR-10 with BadNets and STL-10, and 0.7499 with GTSRB and STL-10) even with a single shadow model and no reserved clean samples. The auxiliary design with a similar MNTD structure further improves performance.

%% file: sec/5_experiment.tex
\section{Experiments}\label{sec: Experiments}
We overview the experimental setup, including datasets, model architectures, attack methods, and defense baselines, consistent with recent works~\citep{ct, scaleup, onepixel, teco}. We then present the experimental results and hyperparameter study.

\subsection{Experimental Setup}\label{sec:Experimental Setup}

\paragraph{Datasets and Model Architectures.}
We use five image datasets: CIFAR-10~\citep{Krizhevsky2009LearningML}, GTSRB~\citep{Stallkamp-IJCNN-2011}, and STL-10~\citep{coates2011stl10}, Tiny-ImageNet~\citep{le2015tiny}, and ImageNet~\citep{ILSVRC15}. 
For a suspicious model $f_S(\cdot)$ trained on CIFAR-10, GTSRB, Tiny-ImageNet or ImageNet, we first train shadow models $f_i$'s using an $\alpha\%$ ($\alpha\in \{1, 5, 10\}$) subset of the corresponding test set as $D_S$. 
Then, we apply VP on $f_i$'s using STL-10 as $D_T$ to obtain the corresponding prompted models $\tilde{f}_i$'s. 
We experiment with ResNet18 and MobileNetV2 architectures, training models on each $D^i_S$ and $D_T^{\text{train}}$ using standard procedures. For the meta-classifier $f_{\text{meta}}$, we use a random forest with 10,000 trees to detect backdoors based on confidence vectors. We mainly use Area Under the ROC Curve (AUROC) and F1-score to measure the detection effectiveness of backdoor detection methods. Our experiments were performed on a workstation equipped with a 16-core Intel i9 CPU (64GB RAM) and an RTX4090 GPU.


\paragraph{Attack Methods and Defense Baselines.}
We evaluate \textsc{BProm} against 9 backdoor poisoning attacks from the Backdoor Toolbox\footnote{\url{https://github.com/vtu81/backdoor-toolbox}}, including classical dirty label, clean label, sample-specific trigger, and adaptive attacks. Default hyperparameters are used to ensure at least 98\% attack success rate.
We compare \textsc{BProm} with 10 backdoor defenses either from Backdoor Toolbox or from their official code. Default hyperparameters are used for each defense. 

\subsection{Experimental Result}
We know from class subspace inconsistency that a prompted model's accuracy degrades if the suspicious model is backdoored. We conducted experiments with backdoor attacks using varying trigger sizes ($4\times 4$, $8\times 
 8$, $16\times 16$ pixels) and poisoning rates (5\%, 10\%, 20\% of training data) to further examine the impact of class subspace inconsistency on prompted model accuracy. For each experiment, we generated a backdoor-infected model and prompted it for a new task on STL-10.  These experiments also cover adaptive attacks, where \textsc{BProm} maintains high performance, achieving an AUROC of 1 even at low poison rates (e.g., 0.2\% for BadNets on CIFAR-10; see Section~\ref{sec:Adaptive Attack}).

\input{tbls/trigger_size}

\textbf{Trigger Size Impact.} Table~\ref{tab:trigger_size} shows the accuracy of prompted models on STL-10 with varying trigger sizes. We trained backdoored models on CIFAR-10 and GTSRB, then prompted them to classify STL-10. As trigger size increases, accuracy decreases. This is because larger triggers distort feature representations more, worsening class subspace inconsistency. 

\input{tbls/poison_rate}

\textbf{Poison Rate Impact.} Table~\ref{tab:poison_rate} shows the accuracy of prompted models with varying poison rates. Similar to the trigger size experiments, we trained backdoored models on CIFAR-10 and GTSRB, then prompted them for STL-10. Higher poison rates lead to lower accuracy due to increased feature distortion, consistent with our class subspace inconsistency explanation. Both Table~\ref{tab:trigger_size} and Table~\ref{tab:poison_rate} show low accuracies, supporting this reasoning.

\textbf{Performance on CIFAR-10 and GTSRB Baselines.} Table~\ref{tab:resnet_defenses_auroc} compares defenses using ResNet18 as the shadow and suspicious model (infected ResNet18 has accuracy $>0.92$ and attack success rate (ASR) $>0.98$, shown in Table~\ref{tab:resnet18_acc_asr} of Section~\ref{sec: Accuracy and Attack Success Rate}). The meta-classifier, trained on Badnets-infected shadow models, classifies suspicious models under 9 attacks. Results from 30 clean and 30 backdoored suspicious models (Section~\ref{sec:Experimental Setup}) show \textsc{BProm} outperforms all other defenses in average AUROC, even when the attack differs from the one used to train the meta-classifier.

\textsc{BProm} achieves high AUROC using only 10\% of the CIFAR-10 (GTSRB) test dataset as the reserved clean dataset $D_S$. Please see Table~\ref{tab:bprom_variations} in Section~\ref{sec: AUROC and F1 Score} for \textsc{BProm} (5\%)'s and \textsc{BProm} (1\%)'s results. In contrast, baseline defenses' AUROC varies significantly across attacks and is heavily influenced by backdoor type. Defenses using activations or saliency maps fail against invisible backdoors spread throughout the image~\citep{Qi2023RevisitingTA}, while perturbation and frequency-based methods cannot handle sample-specific or randomized triggers~\citep{Nguyen2021WaNetI,input-aware}. Tables~\ref{tab:mobilenet_defenses_auroc} and \ref{tab:mobilenet_defenses_f1} in Section~\ref{sec: AUROC and F1 Score} show that \textsc{BProm} maintains high AUROC even with different architectures like MobileNetV2. We also evaluate \textsc{BProm} on MobileViT and Swim Transformer, demonstrating its effectiveness across different architectures (see Section~\ref{sec:architecture_variations} for details). We also tested feature-based backdoors like Refool~\citep{refool}, BPP~\citep{bpp}, and Poison Ink~\citep{poisonink}, with results in Table~\ref{sec: refool-bpp-poisonink} of Section~\ref{sec: AUROC and F1 Score} showing perfect detection. 

\input{tbls/resnet_defenses_auroc}

\input{tbls/tiny-imagenet-auc}

\textbf{Performance on Tiny-ImageNet and ImageNet.} In addition to the CIFAR-10 and GTSRB datasets, we also evaluated \textsc{BProm} on the Tiny-ImageNet and ImageNet datasets. These larger datasets present greater challenges for backdoor detection due to the increased complexity of the images and the larger number of classes. 
Table~\ref{tab:tinyimagenet_auc} (Table~\ref{tab:imagenet_auc} in Section~\ref{sec: Experimental Results on ImageNet}) shows the results on Tiny-ImageNet (ImageNet), comparing \textsc{BProm} with several state-of-the-art defenses. In particular, for Tiny-ImageNet, \textsc{BProm} achieves an average AUROC of 0.899 for ResNet18 and 0.912 for MobileNet, significantly outperforming other defenses.

\textbf{Training Time of \textsc{BProm}.} \textsc{BProm}'s training time, while longer due to shadow model and meta-classifier training, remains practical for deployment given its accuracy and black-box nature. \textsc{BProm}'s training time with different shadow model counts and architectures (CIFAR-10 as $D_S$, STL-10 as $D_T$) is shown below. In particular, for ResNet18, \textsc{BProm}'s training time is 2.3, 4.8, and 9.5 hours if 10, 20, 40 shadow models are considered, respectively. For MobileNetV2, \textsc{BProm}'s training time is 1.2, 2.4, and 5.2 hours if 10, 20, 40 shadow models are considered, respectively. Reported times are averaged over five trials.

\subsection{Hyperparameter Study}
We conduct hyperparameter studies to analyze key factors affecting \textsc{BProm}'s effectiveness.

\input{tbls/shadowing_results}

\paragraph{Impact of Number of Shadow Models.} 
Table~\ref{tab:shadowing_results} shows AUROC as we vary the number of shadow models used to train the backdoor classifier. In the table, “2 (1+1)” means one clean and one backdoor shadow model. The F1 score increases rapidly with more shadow models but plateaus after about 20 models. This indicates that approximately 20 shadow models are sufficient for effective training, with minimal AUROC improvement beyond this number.

\input{tbls/asr_roc_triger_size}

\paragraph{Impact of Trigger Size and Poison Rate.} We analyze how detection performance (AUROC) changes with varying trigger size and poison rate. The settings in Tables~\ref{tab:trigger_metrics} and \ref{tab:poison_metrics} match those in Tables~\ref{tab:trigger_size} and \ref{tab:poison_rate}, which show the prompted model accuracy for different trigger sizes and poison rates. 
Tables~\ref{tab:trigger_metrics} and \ref{tab:poison_metrics} show both attack success rate (ASR) and AUROC for CIFAR-10 models as trigger size and poison rate vary.

\input{tbls/asr_roc_poison_rate}

We observe two key points: 1) ASR increases with larger trigger sizes and poison rates, indicating stronger backdoor attacks. 2) Despite stronger attacks, our detection method's AUROC remains stable, with minor fluctuations. GTSRB results show similar trends: as trigger size increases from 4×4 to 16×16, ASR rises from 26\% to 99\%, while AUROC stays between 0.98 and 1.00. This demonstrates that our backdoor detection technique remains reliable even as attacks strengthen, highlighting its robustness against varying attack strengths.

\paragraph{Structural Differences between Shadow and Suspicious Models.} We analyze the impact of using different architectures for shadow and suspicious models on \textsc{BProm}'s performance. Table~\ref{tab:structure_diff} shows AUROC results with MobileNetV2 as the suspicious model and ResNet18 as the shadow model, indicating that \textsc{BProm}'s detection effectiveness remains robust despite structural differences.

\input{tbls/structure_diff}

\paragraph{Impact of External Dataset.}
We ran additional experiments with $D_S$ as CIFAR-10/GTSRB and $D_T$ changed to SVHN. Results in Tables~\ref{tab: gtsrb-svhn} and \ref{tab: cifar10-svhn} of Section~\ref{sec: AUROC and F1 Score} show consistent detection performance.

\paragraph{Impact of the Inconsistency between Numbers of classes in $D_S$ and $D_T$.}
In previous experiments, we used CIFAR-10 and GTSRB as $D_S$ and STL-10 as $D_T$, maintaining class consistency between $D_S$ and $D_T$. We also ran experiments with $D_T$ as STL-10 and $D_S$ as CIFAR-100. The results in Table~\ref{tab: cifar-100 auroc} of Section~\ref{sec: AUROC and F1 Score} still show consistent detection performance.

\begin{wraptable}{r}{7cm}
\centering
\vspace{-0.5cm}
\caption{Adaptive attacks with low poison rate.}
\label{tab:adaptive_attack_results}
\begin{adjustbox}{max width=0.49\columnwidth}
\begin{tabular}{ccc|ccc}
\hline
Poison Rate & AUROC & ASR &Poison Rate & AUROC & ASR \\
\hline
0.2\% & 1 & 0.709 & 2\% & 1 & 1\\
0.5\% & 1 & 0.838 & 5\% & 1 & 1\\
1\% & 1 & 1 & 10\% & 1 & 1\\
\hline
\end{tabular}
\end{adjustbox}
\vspace{-0.2cm}
\end{wraptable}
\subsection{Adaptive Attack}\label{sec:Adaptive Attack}
To evaluate \textsc{BProm}'s robustness against adaptive attacks, we followed the experimental setup described in \citet{scaleup} (Section 5.3.2), focusing on BadNets attacks on CIFAR-10. It remains unknown how an attacker adds a regularization term to reduce class subspace inconsistency. We examine two candidate adaptive attacks below. 

First, as shown in \cite{revisitingbackdoor}, the backdoor with a very low poison rate can act as an adaptive attack. Table~\ref{tab:adaptive_attack_results} presents the AUROC and ASR of \textsc{BProm} at various poison rates. These results show that \textsc{BProm} maintains perfect detection (AUROC = 1) even at extremely low poison rates, demonstrating its effectiveness against stealthy adaptive attacks.  Our observed ASR values for BadNets at 0.2\% and 0.5\% poison rates align with those reported in Figure 7b of \citet{scaleup}, validating the correctness of our implementation.

\begin{wraptable}{r}{7cm}
\centering
\vspace{-0.5cm}
\caption{Adaptive attacks with clean labels.}
\label{tab:clean_label_attacks}
\begin{adjustbox}{max width=0.6\columnwidth}
\begin{tabular}{lcc}
\hline
Dataset & SIG & LC \\
\hline
CIFAR-10 & 1.00 & 0.95 \\
GTSRB & 0.83 & 0.78 \\
\hline
\end{tabular}
\end{adjustbox}
\vspace{-0.3cm}
\end{wraptable}

Clean-label backdoors, like SIG~\citep{8802997} and LC~\citep{lc} can also be regarded as a different adaptive attack. These attacks do not modify labels and only poison a portion of the training images, potentially preserving class subspaces and hindering \textsc{BProm}'s detection based on class subspace inconsistency.  \textsc{BProm}.  Table~\ref{tab:clean_label_attacks} shows \textsc{BProm}'s performance on SIG and LC. While not perfect, \textsc{BProm} still achieves decent AUROC, indicating its resilience even against these challenging attacks.


%% file: tbls/trigger_size.tex
\begin{wraptable}{r}{7cm}
\centering
\tiny 

\vspace{-0.5cm}
\caption{Testing accuracy for different trigger sizes.}
\label{tab:trigger_size}
\begin{adjustbox}{max width=0.5\columnwidth}
\begin{tabular}{lcccc}
\toprule
\multirow{2}{*}{} & \multicolumn{2}{c}{CIFAR-10} & \multicolumn{2}{c}{GTSRB} \\
\cmidrule(lr){2-3} \cmidrule(lr){4-5}
&\parbox{0.9cm}{ \tiny Blend} & \parbox{0.9cm}{ \tiny Adap-Blend} & \parbox{0.9cm}{ \tiny Blend} & \parbox{0.9cm}{ \tiny Adap-Blend} \\

\midrule
(4*4) & 0.3830 & 0.3336 & 0.1783 & 0.1245 \\
(8*8) & 0.3517 & 0.3250 & 0.1641 & 0.1183 \\
(16*16) & 0.3172 & 0.3127 & 0.1571 & 0.1080 \\
\bottomrule
\end{tabular}
\end{adjustbox}
\vspace{-0.5cm}

\end{wraptable}

%% file: tbls/poison_rate.tex
\begin{wraptable}{r}{7cm}
\centering
\tiny
\vspace{-0.1cm}
\caption{Testing accuracy for various poison rates.}
\label{tab:poison_rate}
\begin{adjustbox}{max width=0.5\columnwidth}
\begin{tabular}{ccccc}
\toprule
\multirow{2}{*}{} & \multicolumn{2}{c}{CIFAR-10} & \multicolumn{2}{c}{GTSRB} \\
\cmidrule(lr){2-3} \cmidrule(lr){4-5}
&\parbox{0.9cm}{ \tiny Blend} & \parbox{0.9cm}{ \tiny Adap-Blend} & \parbox{0.9cm}{ \tiny Blend} & \parbox{0.9cm}{ \tiny Adap-Blend} \\
\midrule
5\% & 0.5297 & 0.5233 & 0.2488 & 0.2368 \\
10\% & 0.4772 & 0.4830 & 0.2328 & 0.2036 \\
20\% & 0.3985 & 0.3358 & 0.2222 & 0.1705 \\
\bottomrule
\end{tabular}
\end{adjustbox}
\vspace{-.1cm}
\end{wraptable}

%% file: tbls/resnet_defenses_auroc.tex
\begin{table*}[t]
\centering
\scriptsize

\captionof{table}{Area Under the ROC Curve (AUROC) of defenses on ResNet18 with different datasets. AVG stands for the average AUROC. Green (red) cells denote values greater (lower) than 0.8.}
\begin{adjustbox}{max width=1.2\columnwidth}
\begin{tabularx}{\textwidth}{lXXXXXXXXXX}
\toprule
\parbox{0.0cm}{} & \parbox{0.0cm}{} & \parbox{0.9cm}{\tiny Badnets\\~\citep{Gu2017BadNetsIV}} & \parbox{0.9cm}{\tiny Blend\\~\citep{Chen2017TargetedBA}} & \parbox{0.9cm}{\tiny Trojan\\~\citep{Liu2018TrojaningAO}} & \parbox{0.5cm}{\tiny BPP\\~\citep{bpp}} & \parbox{1.0cm}{\tiny WaNet\\~\citep{Nguyen2021WaNetI}} & \parbox{0.9cm}{\tiny Dynamic\\~\citep{input-aware}} & \parbox{1.0cm}{\tiny Adap-\\Blend\\~\citep{revisitingbackdoor}} & \parbox{0.8cm}{\tiny Adap-\\Patch\\~\citep{revisitingbackdoor}} & \parbox{0.5cm}{\tiny AVG} \\
\midrule
\multirow{2}{*}{\tiny STRIP~\citep{Gao2019STRIPAD}} & cifar10 & \cellcolor{green!25}0.937 & \cellcolor{green!25}0.834 & \cellcolor{red!25}0.517 & \cellcolor{red!25}0.499 & \cellcolor{red!25}0.499 & \cellcolor{green!25}0.955 & \cellcolor{red!25}0.787 & \cellcolor{red!25}0.520 & \cellcolor{red!25}0.694\\
& gtsrb & \cellcolor{green!25}0.955 & \cellcolor{red!25}0.772 & \cellcolor{red!25}0.670 & \cellcolor{red!25}0.500 & \cellcolor{red!25}0.500 & \cellcolor{green!25}0.971 & \cellcolor{green!25}0.917 & \cellcolor{red!25}0.577 & \cellcolor{red!25}0.733\\
\midrule
\multirow{2}{*}{\shortstack{\tiny AC~\citep{Chen2018DetectingBA}}} & cifar10 & \cellcolor{green!25}0.999 & \cellcolor{green!25}0.992 & \cellcolor{green!25}1.000 & \cellcolor{red!25}0.500 & \cellcolor{red!25}0.500 & \cellcolor{green!25}0.958 & \cellcolor{green!25}0.958 & \cellcolor{green!25}1.000 & \cellcolor{green!25}0.863\\
& gtsrb & \cellcolor{red!25}0.322 & \cellcolor{red!25}0.435 & \cellcolor{red!25}0.255 & \cellcolor{red!25}0.501 & \cellcolor{red!25}0.501 & \cellcolor{red!25}0.696 & \cellcolor{red!25}0.694 & \cellcolor{red!25}0.787 & \cellcolor{red!25}0.524\\
\midrule
\multirow{2}{*}{\tiny Frequency~\citep{low-frequency}} & cifar10 & \cellcolor{green!25}1.000 & \cellcolor{green!25}0.936 & \cellcolor{green!25}1.000 & \cellcolor{green!25}0.999 & \cellcolor{green!25}0.999 & \cellcolor{green!25}0.969 & \cellcolor{green!25}0.896 & \cellcolor{green!25}0.902 & \cellcolor{green!25}0.963\\
& gtsrb & \cellcolor{green!25}0.999 & \cellcolor{green!25}0.939 & \cellcolor{green!25}0.999 & \cellcolor{green!25}0.998 & \cellcolor{green!25}0.998 & \cellcolor{green!25}0.959 & \cellcolor{green!25}0.832 & \cellcolor{green!25}0.879 & \cellcolor{green!25}0.950\\
\midrule
\multirow{2}{*}{\tiny SentiNet~\citep{Chou2018SentiNetDL}} & cifar10 & \cellcolor{green!25}0.949 & \cellcolor{red!25}0.463 & \cellcolor{green!25}0.949 & \cellcolor{red!25}0.502 & \cellcolor{red!25}0.502 & \cellcolor{green!25}0.949 & \cellcolor{red!25}0.470 & \cellcolor{green!25}0.947 & \cellcolor{red!25}0.716\\
& gtsrb & \cellcolor{green!25}0.949 & \cellcolor{red!25}0.590 & \cellcolor{green!25}0.949 & \cellcolor{red!25}0.503 & \cellcolor{red!25}0.503 & \cellcolor{green!25}0.949 & \cellcolor{green!25}0.814 & \cellcolor{green!25}0.949 & \cellcolor{red!25}0.776\\
\midrule
\multirow{2}{*}{\tiny CT~\citep{ct}} & cifar10 & \cellcolor{green!25}0.9898 & \cellcolor{green!25}0.921 & \cellcolor{green!25}0.999 & \cellcolor{red!25}0.502 & \cellcolor{red!25}0.502 & \cellcolor{green!25}0.991 & \cellcolor{green!25}0.954 & \cellcolor{green!25}0.859 & \cellcolor{green!25}0.840\\
& gtsrb & \cellcolor{green!25}0.967 & \cellcolor{green!25}0.978 & \cellcolor{green!25}0.999 & \cellcolor{red!25}0.504 & \cellcolor{red!25}0.504 & \cellcolor{green!25}0.955 & \cellcolor{green!25}0.983 & \cellcolor{green!25}0.861 & \cellcolor{green!25}0.844\\
\midrule
\multirow{2}{*}{\tiny SS~\citet{Tran2018SpectralSI}} & cifar10 & \cellcolor{green!25}0.929 & \cellcolor{green!25}0.921 & \cellcolor{red!25}0.446 & \cellcolor{red!25}0.503 & \cellcolor{red!25}0.503 & \cellcolor{green!25}0.920 & \cellcolor{green!25}0.926 & \cellcolor{green!25}0.830 & \cellcolor{red!25}0.747\\
& gtsrb & \cellcolor{green!25}0.808 & \cellcolor{red!25}0.722 & \cellcolor{green!25}0.800 & \cellcolor{red!25}0.502 & \cellcolor{red!25}0.502 & \cellcolor{green!25}0.800 & \cellcolor{red!25}0.722 & \cellcolor{red!25}0.680 & \cellcolor{red!25}0.692\\
\midrule
\multirow{2}{*}{\tiny SCAn~\citep{scan}} & cifar10 & \cellcolor{green!25}0.985 & \cellcolor{green!25}0.983 & \cellcolor{green!25}0.986 & \cellcolor{red!25}0.498 & \cellcolor{red!25}0.498 & \cellcolor{green!25}0.991 & \cellcolor{green!25}0.815 & \cellcolor{green!25}0.819 & \cellcolor{green!25}0.822\\
& gtsrb & \cellcolor{green!25}0.994 & \cellcolor{green!25}0.956 & \cellcolor{green!25}1.000 & \cellcolor{red!25}0.500 & \cellcolor{red!25}0.500 & \cellcolor{green!25}0.968 & \cellcolor{green!25}0.845 & \cellcolor{green!25}0.867 & \cellcolor{green!25}0.829\\
\midrule
\multirow{2}{*}{\tiny SPECTRE~\citep{SPECTRE}} & cifar10 & \cellcolor{green!25}0.895 & \cellcolor{red!25}0.765 & \cellcolor{green!25}0.931 & \cellcolor{red!25}0.545 & \cellcolor{red!25}0.545 & \cellcolor{green!25}0.841 & \cellcolor{red!25}0.5123 & \cellcolor{red!25}0.396 & \cellcolor{red!25}0.679\\
& gtsrb & \cellcolor{green!25}0.911 & \cellcolor{red!25}0.599 & \cellcolor{green!25}0.800 & \cellcolor{red!25}0.502 & \cellcolor{red!25}0.502 & \cellcolor{red!25}0.567 & \cellcolor{red!25}0.615 & \cellcolor{red!25}0.626 & \cellcolor{red!25}0.640\\
\midrule
\multirow{2}{*}{\tiny MM-BD~\citep{mmbd}} & cifar10 & \cellcolor{green!25}0.867 & \cellcolor{red!25}0.633 & \cellcolor{green!25}0.867 & \cellcolor{green!25}0.867 & \cellcolor{green!25}0.867 & \cellcolor{green!25}0.867 & \cellcolor{green!25}0.867 & \cellcolor{green!25}0.867 & \cellcolor{green!25}0.838\\
& gtsrb & \cellcolor{red!25}0.567 & \cellcolor{red!25}0.633 & \cellcolor{red!25}0.500 & \cellcolor{red!25}0.633 & \cellcolor{red!25}0.767 & \cellcolor{red!25}0.567 & \cellcolor{green!25}0.833 & \cellcolor{green!25}0.833 & \cellcolor{red!25}0.667\\
\midrule
\multirow{2}{*}{\tiny TED~\citep{ted}} & cifar10 & \cellcolor{red!25}0.642 & \cellcolor{red!25}0.485 & \cellcolor{red!25}0.503 & \cellcolor{red!25}0.411 & \cellcolor{red!25}0.676 & \cellcolor{red!25}0.433 & \cellcolor{red!25}0.526 & \cellcolor{red!25}0.664 & \cellcolor{red!25}0.543\\
& gtsrb & \cellcolor{green!25}0.842 & \cellcolor{green!25}0.843 & \cellcolor{red!25}0.558 & \cellcolor{red!25}0.589 & \cellcolor{red!25}0.501 & \cellcolor{red!25}0.663 & \cellcolor{green!25}0.885 & \cellcolor{green!25}0.864 & \cellcolor{red!25}0.718\\
\midrule
\multirow{2}{*}{\shortstack{\tiny \textsc{BProm} (10\%)}} & cifar10 & \cellcolor{green!25}1.000 & \cellcolor{green!25}1.000 & \cellcolor{green!25}1.000 & \cellcolor{green!25}1.000 & \cellcolor{green!25}1.000 & \cellcolor{green!25}1.000 & \cellcolor{green!25}1.000 & \cellcolor{green!25}1.000 & \cellcolor{green!25}1.000\\
& gtsrb & \cellcolor{green!25}1.000 & \cellcolor{green!25}1.000 & \cellcolor{green!25}1.000 & \cellcolor{green!25}0.933 & \cellcolor{green!25}0.933 & \cellcolor{green!25}1.000 & \cellcolor{green!25}1.000 & \cellcolor{green!25}1.000 & \cellcolor{green!25}0.983\\
\bottomrule
\end{tabularx}
\end{adjustbox}
\label{tab:resnet_defenses_auroc}
\vspace{-0.2cm}
\end{table*}

%% file: tbls/tiny-imagenet-auc.tex
\begin{table}[ht]
\centering
\scriptsize
\vspace{-0.3cm}
\captionof{table}{AUROC of defenses on Tiny-ImageNet, using ResNet18 and MobileNetV2. AVG stands for the average AUROC. Green (red) cells denote values greater (lower) than 0.8.}
\begin{adjustbox}{max width=1\columnwidth}
\begin{tabularx}{\textwidth}{l cXXXXXXXX}
\toprule
 &  & \parbox{0.8cm}{Badnets} & \parbox{0.8cm}{Blend} & \parbox{0.8cm}{Trojan} & \parbox{0.5cm}{BPP} & \parbox{0.8cm}{\tiny WaNet} & \parbox{0.8cm}{Adap-\\Blend} & \parbox{0.8cm}{Adap-\\Patch} & \parbox{0.5cm}{AVG} \\
\midrule

\multirow{2}{*}{\tiny STRIP~\citep{Gao2019STRIPAD}} & \tiny ResNet18 & \cellcolor{green!25}0.938 & \cellcolor{green!25}0.905  & \cellcolor{red!25}0.440 & \cellcolor{red!25}0.500 & \cellcolor{red!25}0.500 & \cellcolor{green!25}0.914 & \cellcolor{green!25}0.925 &  \cellcolor{red!25}0.732 \\
\cmidrule{2-10}
& \tiny MobileNetV2 & \cellcolor{green!25}0.936 & \cellcolor{green!25}0.935 & \cellcolor{green!25}0.940 & \cellcolor{red!25}0.500 & \cellcolor{red!25}0.500 & \cellcolor{green!25}0.838 & \cellcolor{green!25}0.830 &  \cellcolor{red!25}0.783 \\
\midrule
\multirow{2}{*}{\tiny AC~\citep{Chen2018DetectingBA}} & \tiny ResNet18 & \cellcolor{red!25}0.490 & \cellcolor{red!25}0.475 & \cellcolor{red!25}0.473 & \cellcolor{red!25}0.501 & \cellcolor{red!25}0.501 & \cellcolor{red!25}0.492 & \cellcolor{red!25}0.491 &  \cellcolor{red!25}0.489 \\
\cmidrule{2-10}
& \tiny MobileNetV2 & \cellcolor{red!25}0.489 &  \cellcolor{red!25}0.485 & \cellcolor{red!25}0.480 & \cellcolor{red!25}0.500 & \cellcolor{red!25}0.500 & \cellcolor{red!25}0.617 & \cellcolor{red!25}0.487 &  \cellcolor{red!25}0.508 \\
\midrule
\multirow{2}{*}{\tiny SS~\citet{Tran2018SpectralSI}} & \tiny ResNet18 & \cellcolor{red!25}0.505 & \cellcolor{red!25}0.485 & \cellcolor{red!25}0.488 & \cellcolor{red!25}0.499 & \cellcolor{red!25}0.499 & \cellcolor{red!25}0.487 & \cellcolor{red!25}0.502 &  \cellcolor{red!25}0.495 \\
\cmidrule{2-10}
& \tiny MobileNetV2 & \cellcolor{red!25}0.487 & \cellcolor{red!25}0.486 & \cellcolor{red!25}0.487 & \cellcolor{red!25}0.502 & \cellcolor{red!25}0.502 & \cellcolor{red!25}0.488 & \cellcolor{red!25}0.500 &  \cellcolor{red!25}0.493 \\
\midrule
\multirow{2}{*}{\tiny SCAn~\citep{scan}} & \tiny ResNet18 & \cellcolor{green!25}0.987 & \cellcolor{green!25}0.987 & \cellcolor{green!25}0.994 & \cellcolor{red!25}0.502 & \cellcolor{red!25}0.502 & \cellcolor{red!25}0.741 & \cellcolor{red!25}0.788 &  \cellcolor{red!25}0.786 \\
\cmidrule{2-10}
& \tiny MobileNetV2 & \cellcolor{green!25}0.982 & \cellcolor{green!25}0.987 & \cellcolor{green!25}0.986 & \cellcolor{red!25}0.502 & \cellcolor{red!25}0.502 & \cellcolor{green!25}0.888 & \cellcolor{green!25}0.882 &  \cellcolor{green!25}0.818 \\
\midrule
\multirow{2}{*}{\tiny CT~\citep{ct}} & \tiny ResNet18 & \cellcolor{green!25}0.945 & \cellcolor{green!25}0.936 & \cellcolor{green!25}0.882 & \cellcolor{red!25}0.501 & \cellcolor{red!25}0.501 & \cellcolor{red!25}0.778 & \cellcolor{red!25}0.776 &  \cellcolor{red!25}0.760 \\
\cmidrule{2-10}
& \tiny MobileNetV2 & \cellcolor{green!25}0.889 & \cellcolor{green!25}0.864 & \cellcolor{green!25}0.915 & \cellcolor{red!25}0.500 & \cellcolor{red!25}0.500 & \cellcolor{green!25}0.823 & \cellcolor{green!25}0.818 &  \cellcolor{red!25}0.758 \\
\midrule
\multirow{2}{*}{\tiny SCALE-UP~\citep{scaleup}} & \tiny ResNet18 & \cellcolor{red!25}0.742 & \cellcolor{red!25}0.724 & \cellcolor{red!25}0.515 & \cellcolor{green!25}1.000 & \cellcolor{green!25}1.000 & \cellcolor{red!25}0.515 & \cellcolor{red!25}0.606 &  \cellcolor{red!25}0.729 \\
\cmidrule{2-10}
& \tiny MobileNetV2 & \cellcolor{red!25}0.651 & \cellcolor{red!25}0.548  & \cellcolor{red!25}0.510 & \cellcolor{green!25}0.980 & \cellcolor{green!25}0.980 & \cellcolor{red!25}0.510 & \cellcolor{red!25}0.717 &  \cellcolor{red!25}0.699 \\
\midrule
\multirow{2}{*}{\tiny CD~\citep{cd}} & \tiny ResNet18 & \cellcolor{green!25}0.918 & \cellcolor{green!25}0.954 & \cellcolor{green!25}0.961 & \cellcolor{red!25}0.628 & \cellcolor{red!25}0.628 & \cellcolor{red!25}0.542 & \cellcolor{red!25}0.647 &  \cellcolor{red!25}0.754 \\
\cmidrule{2-10}
& \tiny MobileNetV2 & \cellcolor{green!25}0.904 & \cellcolor{green!25}0.985  & \cellcolor{green!25}0.997 & \cellcolor{red!25}0.514 & \cellcolor{red!25}0.514 & \cellcolor{red!25}0.591 & \cellcolor{green!25}0.933 &  \cellcolor{green!25}0.805 \\
\midrule
\multirow{2}{*}{\tiny MM-BD~\citep{mmbd}} & \tiny ResNet18 & \cellcolor{green!25}0.800 & \cellcolor{red!25}0.567 & \cellcolor{red!25}0.467 & \cellcolor{green!25}0.967 & \cellcolor{red!25}0.467 & \cellcolor{green!25}0.867 & \cellcolor{green!25}0.867 &  \cellcolor{red!25}0.715 \\
\cmidrule{2-10}
& \tiny MobileNetV2 & \cellcolor{red!25}0.633 & \cellcolor{red!25}0.500 & \cellcolor{red!25}0.467 & \cellcolor{green!25}1.000 & \cellcolor{red!25}0.700 & \cellcolor{red!25}0.633 & \cellcolor{red!25}0.767 &  \cellcolor{red!25}0.671 \\
\midrule
\multirow{2}{*}{\textsc{BProm} (10\%)} & \tiny ResNet18 & \cellcolor{green!25}1.000 & \cellcolor{green!25}0.984 & \cellcolor{green!25}0.900 & \cellcolor{green!25}1.000 & \cellcolor{green!25}1.000 & \cellcolor{green!25}0.966 & \cellcolor{green!25}1.000 &  \cellcolor{green!25}0.979 \\
\cmidrule{2-10}
& \tiny MobileNetV2 & \cellcolor{green!25}1.000 & \cellcolor{green!25}0.978 & \cellcolor{green!25}0.966 & \cellcolor{green!25}1.000 & \cellcolor{green!25}1.000 & \cellcolor{green!25}1.000 & \cellcolor{green!25}1.000 &  \cellcolor{green!25}0.992 \\
\bottomrule
\end{tabularx}
\end{adjustbox}
\label{tab:tinyimagenet_auc}
\vspace{-0.5cm}
\end{table}

%% file: tbls/shadowing_results.tex
\begin{wraptable}{r}{7cm}

\centering
\tiny
\vspace{-0.5cm}
\caption{AUROC relative to the number of shadow models in meta-classifier training.}
\label{tab:shadowing_results}
\begin{adjustbox}{max width=0.6\columnwidth}
\begin{tabular}{ccccc}
\toprule
 & \multicolumn{2}{c}{CIFAR-10} & \multicolumn{2}{c}{GTSRB} \\
\cmidrule(lr){2-3} \cmidrule(lr){4-5}
 \# Shadow Model &\parbox{0.9cm}{ \tiny Blend} & \parbox{0.9cm}{ \tiny Adap-Blend} & \parbox{0.9cm}{ \tiny Blend} & \parbox{0.9cm}{ \tiny Adap-Blend} \\

\midrule
2 (1+1) & 0.667 & 0.938 & 0.789 & 0.967 \\
10 (5+5) & 0.874 & 0.985 & 0.854 & 0.989 \\
20 (10+10) & 1.000  & 1.000  & 1.000  & 1.000      \\
40 (20+20) & 1.000  & 1.000  & 1.000  & 1.000      \\
\bottomrule
\end{tabular}
\end{adjustbox}
\vspace{-0.6cm}
\end{wraptable}

%% file: tbls/asr_roc_triger_size.tex
\begin{wraptable}{r}{7cm}

\centering
\footnotesize
\vspace{-1.2cm}

\caption{ASR and AUROC for Blend and Adap-Blend attacks across different trigger sizes.}
\label{tab:trigger_metrics}

\begin{adjustbox}{max width=0.5\columnwidth}
\begin{tabular}{lcccccc}
 \toprule
 & \multirow{2}{*}{Trigger Size} & \multicolumn{2}{c}{Blend} & \multicolumn{2}{c}{Adap-Blend} \\
\cmidrule(lr){3-4} \cmidrule(lr){5-6}
 & & ASR & AUROC & ASR & AUROC \\
\midrule
\multirow{3}{*}{CIFAR-10} & (4*4) & 0.269 & 1.000 & 0.016 & 1.000 \\
& (8*8) & 0.974 & 1.000 & 0.049 & 1.000 \\
& (16*16) & 0.994 & 1.000 & 0.963 & 1.000 \\
\midrule
\multirow{3}{*}{GTSRB} & (4*4) & 0.842 & 1.000 & 0.027 & 1.000 \\
& (8*8) & 0.994 & 1.000 & 0.194 & 1.000 \\
& (16*16) & 0.994 & 1.000 & 0.997 & 1.000 \\
\bottomrule
\end{tabular}
\end{adjustbox}
\vspace{-0.5cm}
\end{wraptable}

%% file: tbls/asr_roc_poison_rate.tex
\begin{wraptable}{r}{7cm}

\centering
\footnotesize
\vspace{-0.7cm}
\caption{ASR and AUROC for Blend and Adap-Blend attacks at different poison rates.}
\label{tab:poison_metrics}
\begin{adjustbox}{max width=0.5\columnwidth}
\begin{tabular}{lcccccc}
\toprule
  & \multirow{2}{*}{Poison Rate} & \multicolumn{2}{c}{Blend} & \multicolumn{2}{c}{Adap-Blend} \\
\cmidrule(lr){3-4} \cmidrule(lr){5-6}
& & ASR & AUROC & ASR & AUROC \\
\midrule
\multirow{3}{*}{CIFAR-10} & 5\% & 0.996 & 0.607  & 0.998 & 0.607 \\
& 10\% & 0.990 & 0.933 &  0.998 & 0.909 \\
& 20\% & 0.998 & 1.000 &  1.000 & 1.000 \\
\midrule
\multirow{3}{*}{GTSRB} & 5\% & 0.998 & 1.000  & 1.000 & 1.000 \\
& 10\% & 0.998 & 1.000 &  1.000 & 1.000 \\
& 20\% & 0.991 & 1.000 &  1.000 & 1.000 \\
\bottomrule
\end{tabular}
\end{adjustbox}
\vspace{-0.8cm}
\end{wraptable}

%% file: tbls/structure_diff.tex
\begin{wraptable}{r}{7cm}
\centering
\footnotesize
\vspace{-0.5cm}
\caption{F1 score and AUROC of \textsc{BProm} when the suspicious model is MobileNetV2 and the shadow model is ResNet18.}
\label{tab:structure_diff}
\begin{adjustbox}{max width=0.5\columnwidth}
\begin{tabular}{ccccc}
\toprule
& WaNet & Adap-Blend & Adap-Patch & AVG \\
\midrule
F1 & 1.000 & 1.000 & 1.000 & 1.000 \\
\midrule
AUROC & 1.000 & 1.000 & 1.000 & 1.000 \\
\bottomrule
\end{tabular}
\end{adjustbox}
\vspace{-0.2cm}
\end{wraptable}

%% file: sec/6_conclusion.tex
\section{Conclusion and Limitation}\label{sec: Conclusion and Limitation}
We present \textsc{BProm} as a novel VP-based black-box model-level backdoor detection method. \textsc{BProm} relies on class subspace inconsistency, where the prompted model's accuracy degrades if the source model is backdoored. This inconsistency is common in various backdoor attacks due to feature space distortion from the poisoned dataset. Our experiments show \textsc{BProm} effectively detects all-to-one backdoors. However, it struggles with all-to-all backdoors, as their feature space distortion is more controllable by the attacker. Addressing this limitation is left for future work.


%% file: sec/x_appendix.tex
\clearpage
\section*{Appendix of \textsc{BProm}: Black-Box Model-Level Backdoor Detection via Visual Prompting}

This appendix provides additional details and experimental results supporting our main findings. Section A details the implementation and configurations of the experiments. Section B presents \textsc{BProm}'s evaluation on different model architectures, datasets, and attack settings, including analyses of label mapping, class number inconsistency, and feature-based backdoors. Section C provides additional visualizations of class subspace inconsistency to further illustrate our method.

\section{Implementation Details}
We provide details on the configurations of the experiments used to evaluate \textsc{BProm} and other defenses.

\subsection{Attack Configurations}

The configurations of the baseline attacks used in our experiments are summarized in Table~\ref{tab:attack_hyperparam}. For each attack, we specify parameters related to the backdoor trigger insertion, including poison rate and cover rate.
\begin{itemize}
\item Poison rate: The proportion of training data with the trigger pattern. A higher poison rate increases the attacker's influence on the model's behavior but also raises the detection risk.
\item Cover rate: The proportion of data with the trigger pattern that shares the original label. A higher cover rate makes the trigger pattern more stealthy and consistent with the original data distribution but weakens the attack.
\end{itemize}
All attacks are implemented using the default settings in the Backdoor Toolbox\footnote{\url{https://github.com/vtu81/backdoor-toolbox}}; refer to the code repository for more details.

\input{tbls/attack_hyperparam}

\subsection{Defense Configurations}
The important settings used for baseline defenses in our evaluations are summarized below:

\begin{itemize}
\item \textbf{STRIP}~\citep{Gao2019STRIPAD}: Number of superimposing images = 10; defense false positive rate budget = 10\%.
\item \textbf{AC}~\citep{Chen2018DetectingBA}: Cluster threshold = 35\% of class size.
\item \textbf{Frequency}~\citep{low-frequency}: Predicts samples as poisoned or clean using a pretrained binary classifier.
\item \textbf{SentiNet}~\citep{Chou2018SentiNetDL}: FPR = 5\%, number of high activation pixels = top 15\%.
\item \textbf{CT}~\citep{ct}: Confusion iterations  = 6000; confusion factor = 20.
\item \textbf{SS}~\citep{Tran2018SpectralSI}: Number of removed samples = $\min(1.5 \times |D_{poison}|/|D|, 0.5 \times \text{class size})$.
\item \textbf{SCAn}~\citep{scan}: Threshold for abnormal score = 0.5.
\item \textbf{SPECTRE}~\citep{SPECTRE}: Number of removed samples = $\min(1.5 * |D_{poison}|/|D|, 0.5 \times \text{class size})$ from top 50\% suspicious classes.
\end{itemize}

\section{Evaluations on Different Architectures and Datasets}
To evaluate the effectiveness of \textsc{BProm} on different architectures, we conducted experiments using ResNet~\citep{he2016deep} and MobileNetV2~\citep{sandler2018mobilenetv2} as backbone models. . The models are trained on the CIFAR-10~\citep{Krizhevsky2009LearningML} and GTSRB~\citep{Stallkamp-IJCNN-2011} datasets, attacked with 9 different backdoor attacks, and then defended with state-of-the-art methods. 

\subsection{Accuracy and Attack Success Rate}\label{sec: Accuracy and Attack Success Rate}
We report the clean accuracy (ACC) of the infected models on benign test samples without triggers and the attack success rate (ASR), which indicates the percentage of Trojan inputs successfully predicted as the attacker-specified target class. The results are shown in Table~\ref{tab:resnet18_acc_asr} for ResNet18 and Table~\ref{tab:mobilenetv2_acc_asr} for MobileNetV2.

\input{tbls/resnet_acc_asr}
\input{tbls/mobilenet_acc_asr}

The results presented in Table~\ref{tab:resnet18_acc_asr} and Table~\ref{tab:mobilenetv2_acc_asr} reveal that despite maintaining high clean accuracy, both models exhibit very high attack success rates ($>$98\%) across various attacks when triggers are present. This suggests that the backdoors effectively induce misclassification towards the target label. With the effectiveness of the backdoor attacks established, the subsequent evaluation involves assessing the performance of \textsc{BProm} and other state-of-the-art defense methods in detecting these compromised models.

\subsection{AUROC and F1 Score}\label{sec: AUROC and F1 Score}
We evaluate defense methods in detecting backdoor attacks using AUROC and F1 score metrics. Experiments are conducted on CIFAR-10 and GTSRB datasets using ResNet18 and MobileNetV2 architectures to assess and compare detection effectiveness across different model designs. This allows for determining the robustness and architecture-agnostic capability of techniques.

\paragraph{Experiments on ResNet18.}
From the AUROC results in Table~\ref{tab:resnet_defenses_auroc} and F1 scores in Table~\ref{tab:defenses_f1} of defenses evaluated on the ResNet18 model, we observe that \textsc{BProm} demonstrates competitive or superior detection performance over defenses for the majority of attacks. It also significantly elevates the average AUROC and F1 score over the strongest baselines. Although it exhibits relatively lower scores on two attacks, \textsc{BProm} still demonstrates detection capability on par with or better than other methods.

\input{tbls/resnet_defenses_f1}

\paragraph{Experiments on MobileNetV2.}
We further evaluate the effectiveness of backdoor detection methods when using the MobileNetV2 architecture, which utilizes depth-separable convolutions to build a lightweight model. This represents a different design choice than ResNet, which uses residual connections to train deeper models. As shown in Table~\ref{tab:mobilenet_defenses_auroc} and Table~\ref{tab:mobilenet_defenses_f1}, we observe consistently outstanding detection effectiveness of \textsc{BProm} over defenses. 

\input{tbls/mobilenet_defenses_auroc}

\input{tbls/mobilenet_defenses_f1}

The consistent behavior shows that the effectiveness of \textsc{BProm} in detecting backdoors is preserved irrespective of model complexity and design choices.

\paragraph{Experiments on extra external dataset.}
We ran extra experiments, where $D_S$ is kept as CIFAR-10/GTSRB, but $D_T$ is changed to SVHN. Table~\ref{tab: gtsrb-svhn} shows the results when $D_S$ is GTSRB and Table~\ref{tab: cifar10-svhn} shows the results when $D_S$ is CIFAR-10. Both results demonstrate consistent detection performance of \textsc{BProm} even when using a different external dataset $D_T$. This indicates that the choice of external dataset does not significantly impact \textsc{BProm}'s effectiveness.

\begin{table}[ht]
    \centering
    \caption{$D_T$ is changed to SVHN, $D_S$ is kept as GTSRB.}
    \begin{adjustbox}{max width=\columnwidth}
    \begin{tabular}{ccccccccc}
        \toprule
         \parbox{0.0cm}{} & \parbox{1.0cm}{ \scriptsize Badnets\\~\citep{Gu2017BadNetsIV}} & 
         \parbox{0.8cm}{\scriptsize Blend\\~\citep{Chen2017TargetedBA}} & \parbox{0.8cm}{\scriptsize Trojan\\~\citep{Liu2018TrojaningAO}} & \parbox{0.9cm}{\scriptsize WaNet\\~\citep{Nguyen2021WaNetI}} & \parbox{1.0cm}{\scriptsize Dynamic\\~\citep{input-aware}} & \parbox{0.9cm}{\scriptsize Adap-\\Blend\\~\citep{revisitingbackdoor}} & \parbox{0.8cm}{\scriptsize Adap-\\Patch\\~\citep{revisitingbackdoor}} & \parbox{0.5cm}{\scriptsize AVG} \\
         \midrule
         \scriptsize F1 & 0.882 & 1.000 & 1.000 & 0.667 & 1.000 & 0.937 & 1.000 & 0.927\\
        \midrule 
        \scriptsize AUROC & 0.867 & 1.000 & 1.000 & 0.500 & 1.000 & 0.933 & 1.000 & 0.9001\\
        \bottomrule
    \end{tabular}
    \end{adjustbox}
    \label{tab: gtsrb-svhn}
    \vspace{-0.3cm}
\end{table}

\begin{table}[ht]
\caption{$D_T$ is changed to SVHN, $D_S$ is kept as CIFAR-10.}
\centering
\begin{adjustbox}{max width=\columnwidth}
    \begin{tabular}{ccccccccc}
    \toprule
    \parbox{0.9cm}{} & \parbox{0.9cm}{ \scriptsize Badnets\\~\citep{Gu2017BadNetsIV}} & 
         \parbox{0.7cm}{\scriptsize Blend\\~\citep{Chen2017TargetedBA}} & \parbox{0.8cm}{\scriptsize Trojan\\~\citep{Liu2018TrojaningAO}} & \parbox{0.9cm}{\scriptsize WaNet\\~\citep{Nguyen2021WaNetI}} & \parbox{1.0cm}{\scriptsize Dynamic\\~\citep{input-aware}} & \parbox{0.8cm}{\scriptsize Adap-\\Blend\\~\citep{revisitingbackdoor}} & \parbox{0.8cm}{\scriptsize Adap-\\Patch\\~\citep{revisitingbackdoor}} & \parbox{0.5cm}{\scriptsize AVG} \\
     \midrule
     \scriptsize F1 & 1.000 & 1.000 & 1.000 & 0.967 & 1.000 & 1.000 & 1.000 & 0.995\\
    \midrule 
    \scriptsize AUROC & 1.000 & 1.000 & 1.000 & 0.967 & 1.000 & 1.000 & 1.000 & 0.995\\
    \bottomrule
    \end{tabular}
\end{adjustbox}
\label{tab: cifar10-svhn}
\vspace{-0.3cm}
\end{table}

\paragraph{Experiments on CIFAR-100.}
To investigate the impact of inconsistency between the numbers of classes in $D_S$ and $D_T$, we conducted experiments using CIFAR-100 as $D_S$
and STL-10 as $D_T$. Table~\ref{tab: cifar-100 auroc} shows that \textsc{BProm} achieves high AUROC and F1 scores across various backdoor attacks, demonstrating its robustness even when there is a significant mismatch in the number of classes (100 classes in $D_S$
vs. 10 classes in $D_T$). This suggests that \textsc{BProm} is capable of handling scenarios where the source and target domains have different numbers of classes, making it a versatile detection method.

\begin{table}[ht]
    \centering
    \caption{AUROC of defenses on ResNet18 under backdoor attacks on CIFAR-100. AVG stands for the average AUROC.}
    \begin{adjustbox}{max width=\columnwidth}
    \centering
    \begin{tabular}{cccccccc}
    \toprule
    \parbox{0.0cm}{} & 
    \parbox{0.9cm}{ \scriptsize Badnets\\~\citep{Gu2017BadNetsIV}} & 
    \parbox{0.7cm}{\scriptsize Blend\\~\citep{Chen2017TargetedBA}} & 
    \parbox{0.8cm}{\scriptsize Trojan\\~\citep{Liu2018TrojaningAO}} & 
    \parbox{0.9cm}{\scriptsize WaNet\\~\citep{Nguyen2021WaNetI}} & 
    \parbox{0.9cm}{\scriptsize Adap-\\Blend\\~\citep{revisitingbackdoor}} & 
    \parbox{0.8cm}{\scriptsize Adap-\\Patch\\~\citep{revisitingbackdoor}} & 
    \parbox{0.5cm}{ AVG} \\
    \midrule
    \scriptsize STRIP~\citep{Gao2019STRIPAD}  & \cellcolor{green!25}0.876 & \cellcolor{red!25}0.732 & \cellcolor{red!25}0.762 & \cellcolor{red!25}0.135 & \cellcolor{green!25}0.941 & \cellcolor{green!25}0.964 & \cellcolor{red!25}0.729\\
    \midrule
    \scriptsize AC~\citep{Chen2018DetectingBA}  & \cellcolor{red!25}0.000 & \cellcolor{red!25}0.000 & \cellcolor{red!25}0.000 & \cellcolor{red!25}0.000 & \cellcolor{green!25}0.998 & \cellcolor{green!25}1.000 & \cellcolor{red!25}0.333\\
    \midrule
    \scriptsize Frequency~\citep{low-frequency} & \cellcolor{green!25}0.986 & \cellcolor{green!25}0.896 & \cellcolor{green!25}0.895 & \cellcolor{green!25}0.883 & \cellcolor{green!25}0.865 & \cellcolor{green!25}0.985 & \cellcolor{green!25}0.926\\
    \midrule
    \scriptsize SentiNet~\cite{Chou2018SentiNetDL} & \cellcolor{green!25}0.952 & \cellcolor{red!25}0.047 & \cellcolor{green!25}0.952 & \cellcolor{red!25}0.115 & \cellcolor{red!25}0.240 & \cellcolor{green!25}0.952 & \cellcolor{red!25}0.551\\
    \midrule
    \scriptsize SS~\citep{Tran2018SpectralSI}  & \cellcolor{red!25}0.661 & \cellcolor{red!25}0.005 & \cellcolor{red!25}0.661 & \cellcolor{red!25}0.673 & \cellcolor{red!25}0.633 & \cellcolor{red!25}0.672 & \cellcolor{red!25}0.551\\
    \midrule
    \scriptsize SCAn\citep{scan}  & \cellcolor{green!25}0.992 & \cellcolor{green!25}0.980 & \cellcolor{green!25}0.982 & \cellcolor{red!25}0.612 & \cellcolor{green!25}0.877 & \cellcolor{red!25}0.380 & \cellcolor{green!25}0.804\\
    \midrule
    \scriptsize \textsc{BProm}(10\%)  & \cellcolor{green!25}1.000 & \cellcolor{green!25}1.000 & \cellcolor{green!25}1.000 & \cellcolor{green!25}1.000 & \cellcolor{green!25}1.000 & \cellcolor{green!25}1.000 & \cellcolor{green!25}\textbf{1.000}\\                            
    \bottomrule
    \end{tabular}
    \end{adjustbox}
    \label{tab: cifar-100 auroc}
\end{table}

\paragraph{Experiments on feature-based backdoors.}
We further evaluated \textsc{BProm}'s performance on feature-based backdoors, which manipulate the model's feature representations instead of directly modifying input images. Table~\ref{sec: refool-bpp-poisonink} presents the results of \textsc{BProm} on three feature-based backdoor methods: Refool~\citep{refool}, BPP~\citep{bpp}, and Poison Ink~\citep{poisonink}, using the same configuration as previous experiments. The high F1 scores and AUROC values indicate that \textsc{BProm} effectively detects these feature-based backdoors, demonstrating its versatility in handling diverse backdoor attack strategies.

\begin{table}[ht]
    \centering
    \caption{Feature-based backdoors like Refool, BPP, Poison Ink.}
    \begin{adjustbox}{max width=1.0\columnwidth, center}
    \begin{tabular}{cccc}
    \toprule
      Attack  &   Dataset &   F1 Score &   AUROC \\
    \midrule
      Refool~\citep{refool} &   CIFAR-10 & 1.000 & 1.000 \\
    \midrule
      BPP~\citep{bpp} &   CIFAR-10 & 1.000 & 1.000 \\
    \midrule
      Poison Ink~\citep{poisonink} &   CIFAR-10 & 1.000 & 1.000 \\
    \bottomrule
    \end{tabular}
    \end{adjustbox}
    \label{sec: refool-bpp-poisonink}
    \vspace{-0.3cm}
\end{table}

\paragraph{Impact of Reserved Clean Dataset Size.}
We analyze the impact of the reserved clean dataset size ($D_S$) on \textsc{BProm}'s performance.  As shown in Table~\ref{tab:bprom_variations}, \textsc{BProm} maintains high AUROC across different $D_S$ sizes (1\%, 5\%, and 10\% of the CIFAR-10 and GTSRB test sets).  Even with a limited $D_S$ (1\%), \textsc{BProm} achieves competitive performance, demonstrating its efficiency in leveraging small amounts of clean data. This robustness to $D_S$ size makes \textsc{BProm} practical for real-world scenarios where clean data might be scarce.

\begin{table*}[t]
\centering
\scriptsize
\caption{Detailed AUROC of \textsc{BProm} with varying sizes of the reserved clean dataset ($D_S$).}
\label{tab:bprom_variations}
\begin{adjustbox}{max width=1.0\columnwidth}  
\begin{tabularx}{\textwidth}{lXXXXXXXXX}
\toprule
 \parbox{0.0cm}{} & \parbox{0.0cm}{} & \parbox{0.9cm}{ \tiny Badnets\\~\citep{Gu2017BadNetsIV}} & \parbox{0.9cm}{\tiny Blend\\~\citep{Chen2017TargetedBA}} & \parbox{0.9cm}{\tiny Trojan\\~\citep{Liu2018TrojaningAO}} & \parbox{1.0cm}{\tiny WaNet\\~\citep{Nguyen2021WaNetI}} & \parbox{0.9cm}{\tiny Dynamic\\~\citep{input-aware}} & \parbox{1.0cm}{\tiny Adap-\\Blend\\~\citep{revisitingbackdoor}} & \parbox{0.8cm}{\tiny Adap-\\Patch\\~\citep{revisitingbackdoor}} & \parbox{0.5cm}{ AVG} \\

\midrule
\multirow{2}{*}{\textsc{BProm} (10\%)} & cifar10 &  1.000 & 1.000 & 1.000 & 1.000 & 1.000 & 1.000 & 1.000 & \textbf{1.000}\\
& gtsrb & 1.000 & 1.000 & 1.000 & 1.000 & 1.000 & 1.000 & 1.000 & \textbf{1.000}\\
\midrule
\multirow{2}{*}{\textsc{BProm} (5\%)} & cifar10 & 1.000 & 1.000 & 1.000 & 1.000 & 1.000 & 1.000 & 1.000 & \textbf{1.000}\\
& gtsrb & 1.000 & 1.000 & 1.000 & 1.000 & 1.000 & 1.000 & 1.000 & \textbf{1.000}\\
\midrule
\multirow{2}{*}{\textsc{BProm} (1\%)} & cifar10 & 1.000 & 1.000 & 1.000 & 1.000 & 1.000 & 1.000 & 1.000 & \textbf{1.000}\\
& gtsrb & 1.000 & 1.000 & 1.000 & 1.000 & 1.000 & 1.000 & 1.000 & \textbf{1.000}\\
\bottomrule
\end{tabularx}
\end{adjustbox}
\end{table*}

\subsection{\textsc{BProm} Performance on MobileViT and Swim Transformer }\label{sec:architecture_variations}
To demonstrate \textsc{BProm}'s architecture-agnostic nature, we evaluated its performance on MobileViT and Swim Transformer, models combining CNN and transformer components. Tables~\ref{tab:mobilevit_results} and~\ref{tab:swimtransformer_results} present the AUROC scores on CIFAR-10 and GTSRB across various backdoor attacks. The results show that \textsc{BProm} maintains competitive performance on both MobileViT and Swim Transformer, indicating its effectiveness is not limited to ResNet-based architectures. The average AUROC is calculated for each defense and dataset.

\begin{table*}[t]
\centering
\scriptsize

\captionof{table}{AUROC on MobileViT for CIFAR-10 and GTSRB. AVG stands for the average AUROC. Green (red) cells denote values greater (lower) than 0.8.}
\label{tab:mobilevit_results}
\begin{adjustbox}{max width=1.0\columnwidth} 
\begin{tabularx}{\textwidth}{lXXXXXXXXX}
\toprule
\parbox{0.0cm}{} & \parbox{0.0cm}{} & \parbox{0.9cm}{\tiny Badnets\\~\citep{Gu2017BadNetsIV}} & \parbox{0.9cm}{\tiny Blend\\~\citep{Chen2017TargetedBA}} & \parbox{0.9cm}{\tiny Trojan\\~\citep{Liu2018TrojaningAO}} & \parbox{1.0cm}{\tiny WaNet\\~\citep{Nguyen2021WaNetI}} & \parbox{0.9cm}{\tiny Dynamic\\~\citep{input-aware}} & \parbox{1.0cm}{\tiny Adap-\\Blend\\~\citep{revisitingbackdoor}} & \parbox{0.8cm}{\tiny Adap-\\Patch\\~\citep{revisitingbackdoor}} & \parbox{0.5cm}{ AVG} \\

\midrule
\multirow{2}{*}{\tiny STRIP~\citep{Gao2019STRIPAD}} & cifar10 & \cellcolor{red!25}0.4974 & \cellcolor{red!25}0.7775 & \cellcolor{green!25}0.9495 & \cellcolor{red!25}0.4705 & \cellcolor{green!25}0.9555 & \cellcolor{green!25}0.9497 & \cellcolor{green!25}0.9501 & \cellcolor{red!25}0.7929 \\
& gtsrb & \cellcolor{green!25}0.9140 & \cellcolor{green!25}0.900 & \cellcolor{green!25}0.9497 & \cellcolor{red!25}0.4855 & \cellcolor{green!25}0.9501 & \cellcolor{green!25}0.9447 & \cellcolor{green!25}0.9501 & \cellcolor{green!25}0.8706 \\
\midrule
\multirow{2}{*}{\shortstack{\tiny AC~\citep{Chen2018DetectingBA}}} & cifar10 & \cellcolor{red!25}0.4738 & \cellcolor{red!25}0.7745 & \cellcolor{green!25}1.0000 & \cellcolor{red!25}0.4252 & \cellcolor{green!25}0.8334 & \cellcolor{green!25}0.9930 & \cellcolor{green!25}0.8996 & \cellcolor{red!25}0.7714 \\
& gtsrb & \cellcolor{red!25}0.2198 & \cellcolor{red!25}0.2591 & \cellcolor{green!25}1.0000 & \cellcolor{red!25}0.312 & \cellcolor{red!25}0.6702 & \cellcolor{green!25}0.9999 & \cellcolor{green!25}1.0000 & \cellcolor{red!25}0.6373 \\
\midrule
\multirow{2}{*}{\tiny Frequency~\citep{low-frequency}} & cifar10 & \cellcolor{green!25}1.000 & \cellcolor{green!25}0.9962 & \cellcolor{green!25}1.000 & \cellcolor{green!25}0.969 & \cellcolor{green!25}0.9999 & \cellcolor{green!25}0.9961 & \cellcolor{green!25}0.9643 & \cellcolor{green!25}0.9893 \\
& gtsrb & \cellcolor{green!25}0.9994 & \cellcolor{green!25}0.9732 & \cellcolor{green!25}0.9993 & \cellcolor{red!25}0.7782 & \cellcolor{green!25}0.9601 & \cellcolor{green!25}0.8731 & \cellcolor{green!25}0.8993 & \cellcolor{green!25}0.9261 \\
\midrule
\multirow{2}{*}{\tiny CT~\citep{ct}} & cifar10 & \cellcolor{green!25}0.9941 & \cellcolor{green!25}0.9379 & \cellcolor{green!25}0.9843 & \cellcolor{red!25}0.7475 & \cellcolor{green!25}0.9892 & \cellcolor{green!25}0.9439 & \cellcolor{green!25}0.8815 & \cellcolor{green!25}0.9255 \\
& gtsrb & \cellcolor{green!25}0.9727 & \cellcolor{green!25}0.9718 & \cellcolor{green!25}0.9744 & \cellcolor{green!25}0.9677 & \cellcolor{red!25}0.5652 & \cellcolor{green!25}0.9684 & \cellcolor{green!25}0.9289 & \cellcolor{green!25}0.9070 \\
\midrule
\multirow{2}{*}{\tiny SS~\citet{Tran2018SpectralSI}} & cifar10 & \cellcolor{red!25}0.5002 & \cellcolor{red!25}0.3902 & \cellcolor{red!25}0.3745 & \cellcolor{red!25}0.5145 & \cellcolor{red!25}0.6154 & \cellcolor{red!25}0.3801 & \cellcolor{red!25}0.3977 & \cellcolor{red!25}0.4532 \\
& gtsrb & \cellcolor{red!25}0.4925 & \cellcolor{red!25}0.4987 & \cellcolor{red!25}0.4925 & \cellcolor{red!25}0.4925 & \cellcolor{red!25}0.4966 & \cellcolor{red!25}0.4961 & \cellcolor{red!25}0.4929 & \cellcolor{red!25}0.4945 \\
\midrule
\multirow{2}{*}{\tiny SCAn~\citep{scan}} & cifar10 & \cellcolor{red!25}0.5000 & \cellcolor{red!25}0.5000 & \cellcolor{green!25}0.9999 & \cellcolor{red!25}0.6578 & \cellcolor{red!25}0.5652 & \cellcolor{red!25}0.5000 & \cellcolor{red!25}0.5000 & \cellcolor{red!25}0.6104 \\
& gtsrb & \cellcolor{red!25}0.4771 & \cellcolor{red!25}0.5481 & \cellcolor{red!25}0.4747 & \cellcolor{green!25}0.8439 & \cellcolor{red!25}0.5000 & \cellcolor{green!25}0.8758 & \cellcolor{red!25}0.6902 & \cellcolor{red!25}0.6300 \\
\midrule
\multirow{2}{*}{\tiny SPECTRE~\citep{SPECTRE}} & cifar10 & \cellcolor{red!25}0.4752 & \cellcolor{red!25}0.4752 & \cellcolor{red!25}0.5961 & \cellcolor{red!25}0.4754 & \cellcolor{red!25}0.4754 & \cellcolor{red!25}0.5015 & \cellcolor{red!25}0.4752 & \cellcolor{red!25}0.4963 \\
& gtsrb & \cellcolor{red!25}0.7383 & \cellcolor{red!25}0.4995 & \cellcolor{red!25}0.7383 & \cellcolor{red!25}0.4995 & \cellcolor{red!25}0.4995 & \cellcolor{red!25}0.4995 & \cellcolor{red!25}0.4991 & \cellcolor{red!25}0.5677 \\
\midrule
\multirow{2}{*}{\tiny \textsc{BProm} (10\%)} & cifar10 & \cellcolor{green!25}1.0000 & \cellcolor{green!25}1.0000 & \cellcolor{green!25}0.9667 & \cellcolor{green!25}1.0000 & \cellcolor{green!25}1.0000 & \cellcolor{green!25}1.0000 & \cellcolor{green!25}1.0000 & \cellcolor{green!25}0.9952 \\
& gtsrb & \cellcolor{green!25}1.0000 & \cellcolor{green!25}1.0000 & \cellcolor{green!25}1.0000 & \cellcolor{green!25}1.0000 & \cellcolor{green!25}0.9655 & \cellcolor{green!25}0.9655 & \cellcolor{green!25}1.0000 & \cellcolor{green!25}0.9901 \\
\bottomrule
\end{tabularx}
\end{adjustbox}
\vspace{-0.2cm}
\end{table*}

\begin{table*}[t]
\centering
\scriptsize

\captionof{table}{AUROC on Swim Transformer for CIFAR-10 and GTSRB. AVG stands for the average AUROC. Green (red) cells denote values greater (lower) than 0.8.}
\label{tab:swimtransformer_results}
\begin{adjustbox}{max width=1.0\columnwidth} 
\begin{tabularx}{\textwidth}{lXXXXXXXXX}
\toprule
\parbox{0.0cm}{} & \parbox{0.0cm}{} & \parbox{0.9cm}{\tiny Badnets\\~\citep{Gu2017BadNetsIV}} & \parbox{0.9cm}{\tiny Blend\\~\citep{Chen2017TargetedBA}} & \parbox{0.9cm}{\tiny Trojan\\~\citep{Liu2018TrojaningAO}} & \parbox{1.0cm}{\tiny WaNet\\~\citep{Nguyen2021WaNetI}} & \parbox{0.9cm}{\tiny Dynamic\\~\citep{input-aware}} & \parbox{1.0cm}{\tiny Adap-\\Blend\\~\citep{revisitingbackdoor}} & \parbox{0.8cm}{\tiny Adap-\\Patch\\~\citep{revisitingbackdoor}} & \parbox{0.5cm}{ AVG} \\

\midrule
\multirow{2}{*}{\tiny STRIP~\citep{Gao2019STRIPAD}} & cifar10 & \cellcolor{green!25}0.998 & \cellcolor{green!25}0.9761 & \cellcolor{green!25}0.9386 & \cellcolor{red!25}0.483 & \cellcolor{green!25}0.9794 & \cellcolor{green!25}0.8558 & \cellcolor{green!25}0.8622 & \cellcolor{green!25}0.8704 \\
& gtsrb & \cellcolor{green!25}0.9851 & \cellcolor{red!25}0.7922 & \cellcolor{green!25}0.8036 & \cellcolor{red!25}0.5805 & \cellcolor{green!25}0.9999 & \cellcolor{red!25}0.7494 & \cellcolor{green!25}0.8174 & \cellcolor{green!25}0.8183 \\
\midrule
\multirow{2}{*}{\shortstack{\tiny AC~\citep{Chen2018DetectingBA}}} & cifar10 & \cellcolor{red!25}0.5001 & \cellcolor{red!25}0.5005 & \cellcolor{red!25}0.4999 & \cellcolor{red!25}0.5000 & \cellcolor{red!25}0.5002 & \cellcolor{red!25}0.4998 & \cellcolor{red!25}0.5001 & \cellcolor{red!25}0.5001 \\
& gtsrb & \cellcolor{red!25}0.2198 & \cellcolor{red!25}0.2591 & \cellcolor{red!25}0.5001 & \cellcolor{red!25}0.5012 & \cellcolor{red!25}0.5702 & \cellcolor{red!25}0.4999 & \cellcolor{red!25}0.5001 & \cellcolor{red!25}0.4358 \\
\midrule
\multirow{2}{*}{\tiny Frequency~\citep{low-frequency}} & cifar10 & \cellcolor{green!25}1.0000 & \cellcolor{green!25}0.9563 & \cellcolor{green!25}0.9594 & \cellcolor{red!25}0.5707 & \cellcolor{green!25}0.9999 & \cellcolor{green!25}0.8564 & \cellcolor{green!25}0.8125 & \cellcolor{green!25}0.8793 \\
& gtsrb & \cellcolor{green!25}0.9283 & \cellcolor{green!25}0.8835 & \cellcolor{green!25}0.9194 & \cellcolor{red!25}0.6767 & \cellcolor{green!25}0.8301 & \cellcolor{green!25}0.8531 & \cellcolor{green!25}0.8794 & \cellcolor{green!25}0.8529 \\
\midrule
\multirow{2}{*}{\tiny CT~\citep{ct}} & cifar10 & \cellcolor{green!25}0.868 & \cellcolor{green!25}0.9968 & \cellcolor{green!25}0.9994 & \cellcolor{red!25}0.5142 & \cellcolor{green!25}0.9919 & \cellcolor{green!25}0.8766 & \cellcolor{green!25}0.9758 & \cellcolor{green!25}0.8890 \\
& gtsrb & \cellcolor{green!25}0.9125 & \cellcolor{green!25}0.867 & \cellcolor{green!25}0.9967 & \cellcolor{red!25}0.4279 & \cellcolor{green!25}0.9991 & \cellcolor{green!25}0.8638 & \cellcolor{red!25}0.7455 & \cellcolor{green!25}0.8304 \\
\midrule
\multirow{2}{*}{\tiny SS~\citet{Tran2018SpectralSI}} & cifar10 & \cellcolor{red!25}0.3877 & \cellcolor{red!25}0.3745 & \cellcolor{red!25}0.3753 & \cellcolor{red!25}0.2747 & \cellcolor{red!25}0.3749 & \cellcolor{red!25}0.3749 & \cellcolor{red!25}0.3913 & \cellcolor{red!25}0.3648 \\
& gtsrb & \cellcolor{red!25}0.4961 & \cellcolor{red!25}0.4925 & \cellcolor{red!25}0.4946 & \cellcolor{red!25}0.4987 & \cellcolor{red!25}0.4925 & \cellcolor{red!25}0.4925 & \cellcolor{red!25}0.4925 & \cellcolor{red!25}0.4942 \\
\midrule
\multirow{2}{*}{\tiny SCAn~\citep{scan}} & cifar10 & \cellcolor{green!25}0.9943 & \cellcolor{green!25}0.9943 & \cellcolor{green!25}0.9549 & \cellcolor{red!25}0.7596 & \cellcolor{red!25}0.7495 & \cellcolor{green!25}0.8451 & \cellcolor{red!25}0.6215 & \cellcolor{green!25}0.8456 \\
& gtsrb & \cellcolor{green!25}0.8973 & \cellcolor{red!25}0.6869 & \cellcolor{red!25}0.7661 & \cellcolor{red!25}0.5219 & \cellcolor{green!25}0.8141 & \cellcolor{red!25}0.6451 & \cellcolor{red!25}0.5577 & \cellcolor{red!25}0.6984 \\
\midrule
\multirow{2}{*}{\tiny SPECTRE~\citep{SPECTRE}} & cifar10 & \cellcolor{red!25}0.5964 & \cellcolor{red!25}0.5981 & \cellcolor{red!25}0.5959 & \cellcolor{red!25}0.4751 & \cellcolor{red!25}0.5997 & \cellcolor{red!25}0.4751 & \cellcolor{red!25}0.4752 & \cellcolor{red!25}0.5451 \\
& gtsrb & \cellcolor{red!25}0.7383 & \cellcolor{red!25}0.4911 & \cellcolor{red!25}0.7383 & \cellcolor{red!25}0.4911 & \cellcolor{red!25}0.4995 & \cellcolor{red!25}0.4991 & \cellcolor{red!25}0.4911 & \cellcolor{red!25}0.5641 \\
\midrule
\multirow{2}{*}{\tiny \textsc{BProm} (10\%)} & cifar10 & \cellcolor{green!25}1.0000 & \cellcolor{green!25}1.0000 & \cellcolor{green!25}1.0000 & \cellcolor{green!25}1.0000 & \cellcolor{green!25}0.8000 & \cellcolor{green!25}0.8949 & \cellcolor{green!25}0.8667 & \cellcolor{green!25}0.9374 \\
& gtsrb & \cellcolor{green!25}1.0000 & \cellcolor{green!25}1.0000 & \cellcolor{green!25}1.0000 & \cellcolor{green!25}1.0000 & \cellcolor{green!25}1.0000 & \cellcolor{green!25}0.8000 & \cellcolor{green!25}0.9334 & \cellcolor{green!25}0.9619 \\

\bottomrule
\end{tabularx}
\end{adjustbox}
\label{tab:resnet_defenses_auroc}
\vspace{-0.2cm}
\end{table*}

\section{Another Visualization of Class Subspace Inconsistency}\label{sec: Another Visualization of Class Subspace Inconsistency}
\begin{figure}
     \centering
     \begin{subfigure}[b]{0.45\linewidth}
         \centering
         \includegraphics[width=1.0\linewidth]{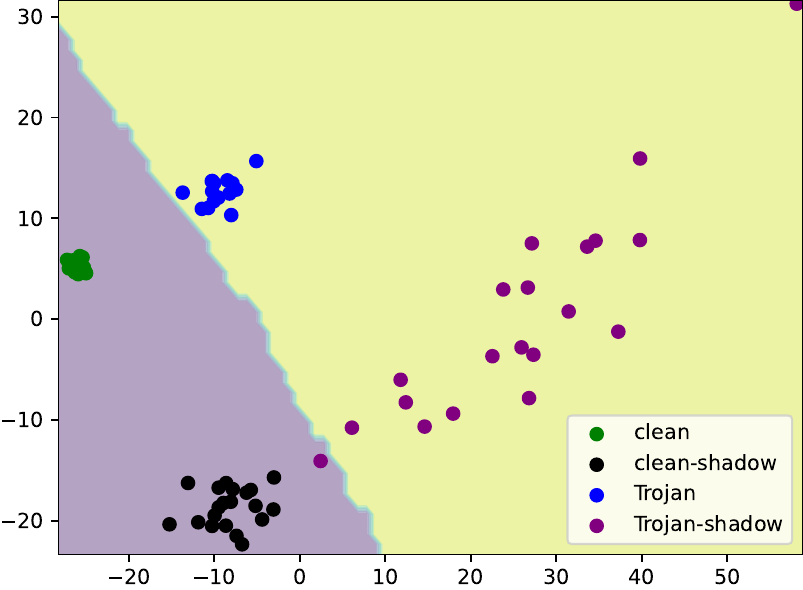}
         \caption{Trojan-infected model.}
         \label{fig: Trojan-infected model}
     \end{subfigure}
     \hfill
     \begin{subfigure}[b]{0.45\linewidth}
         \centering
         \includegraphics[width=1.0\linewidth]{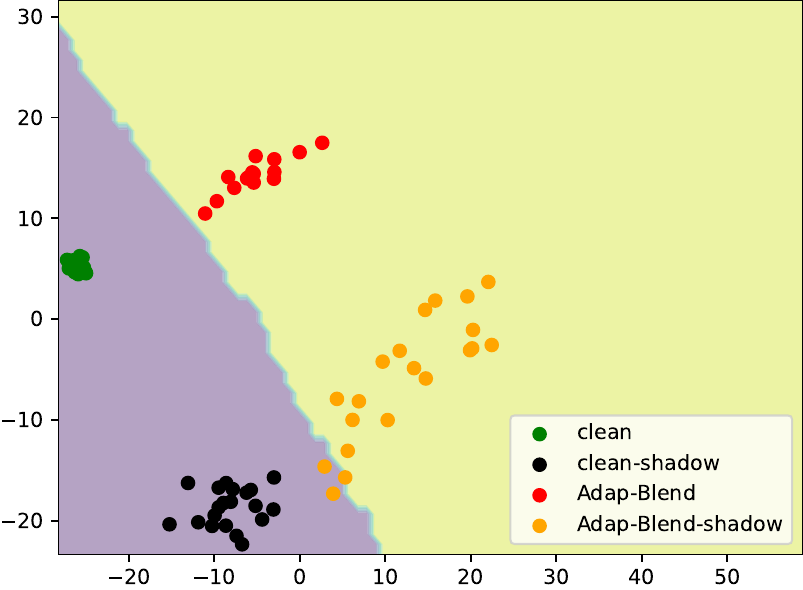}
         \caption{Adaptive-Blend-infected model.}
         \label{fig: Adaptive-Blend-infected model}
     \end{subfigure}
     \caption{Visualization of class subspace inconsistency through PCA.}
     \label{fig:another visualization}
\end{figure}

Figure~\ref{fig: Trojan-infected model} illustrates, using principal component analysis (PCA), 30 suspicious models (15 clean and 15 backdoor) trained on the complete CIFAR-10 dataset, along with 40 shadow models (20 clean-shadow and 20 backdoor-shadow) trained on 10\% of the CIFAR-10 test set. All models are based on ResNet18, with the Trojan method~\cite{Liu2018TrojaningAO} employed as the backdoor technique. Subsequently, a random forest-based meta-model (binary classifier) with 50 estimators is trained on the confidence vectors produced by the 40 shadow models. A distinct separation between clean (\textcolor{green}{green} dots) and backdoor models (\textcolor{blue}{blue} dots) is evident after VP, attributed to class subspace inconsistency. The same meta-model is also used to classify clean (\textcolor{green}{green} dots) and Adap-Blend-infected models (\textcolor{red}{red} dots)~\cite{revisitingbackdoor}. A similar pattern is observable in Figure~\ref{fig: Adaptive-Blend-infected model}.

\section{Experimental Results on ImageNet}\label{sec: Experimental Results on ImageNet}
This section includes the experimental results on ImageNet. In particular, \textsc{BProm} achieves an average AUROC of 0.9996 for ResNet18, significantly outperforming other defenses.
\input{tbls/imagenet_auc}

\section{Notation and Definitions}\label{sec: Notation and Definitions}

For clarity and reproducibility, Table~\ref{tab:notation} summarizes the notation and definitions used throughout the paper.

\begin{table}[h]
    \centering
    \caption{Notation and Definitions}
    \label{tab:notation}
    \begin{tabular}{l|l}
        \hline
        Symbol & Description \\
        \hline
        $D_S$ & Reserved clean dataset (1-10\% of test set) \\
        $D_E$ & Extracted samples from $D_S$ for poisoning \\
        $D_P$ & Poisoned dataset created from $D_S$ \\
        $D_T$ & External clean dataset for visual prompting \\
        $D_Q$ & Random samples from $D_T$'s test set \\
        $D_{\text{meta}}$ & Samples for training meta-classifier \\
        \hline
    \end{tabular}
\end{table}

%% file: tbls/attack_hyperparam.tex




\begin{table}[ht]
\vspace{-0.3cm}
\caption{Configurations of baseline attacks}
\centering
\footnotesize

\begin{adjustbox}{max width=0.75\columnwidth}
\begin{tabularx}{\columnwidth}{XXX}
\toprule
\textbf{Attacks} & \textbf{CIFAR-10} & \textbf{GTSRB} \\
\midrule
BadNets~\cite{Gu2017BadNetsIV} & Poison Rate: 0.3\%  & Poison Rate: 1.0\% \\
\midrule
Blend~\cite{Chen2017TargetedBA} & Poison Rate: 0.3\% & Poison Rate: 1.0\% \\
\midrule
Trojan~\cite{Liu2018TrojaningAO} & Poison Rate: 0.3\% & Poison Rate: 1.0\% \\
\midrule
WaNet~\cite{Nguyen2021WaNetI} & Poison Rate: 5.0\% & Poison Rate: 5.0\% \\
 & Cover Rate: 10.0\% & Cover Rate: 10.0\% \\
\midrule
Dynamic~\cite{input-aware} & Poison Rate: 0.3\% & Poison Rate: 0.3\% \\
\midrule
Adap-Blend~\cite{revisitingbackdoor} & Poison Rate: 0.3\% & Poison Rate: 0.5\% \\
 & Cover Rate: 0.6\% & Cover Rate: 1.0\% \\
\midrule
Adap-Patch~\cite{revisitingbackdoor} & Poison Rate: 0.3\% & Poison Rate: 0.3\% \\
 & Cover Rate: 0.3\% & Cover Rate: 0.6\% \\
\bottomrule
\end{tabularx}
\end{adjustbox}
\label{tab:attack_hyperparam}
\vspace{-0.3cm}
\end{table}

%% file: tbls/resnet_acc_asr.tex
\begin{table}[ht]
\centering
\vspace{-0.3cm}
\captionof{table}{Accuracy and ASR on ResNet18.}
\begin{adjustbox}{max width=1\columnwidth}
\begin{tabularx}{\textwidth}{lXXXXXXXXX}
\toprule
\parbox{0.0cm}{} & \parbox{0.0cm}{} & \parbox{0.9cm}{ \tiny Badnets\\~\cite{Gu2017BadNetsIV}} & \parbox{0.9cm}{\tiny Blend\\~\cite{Chen2017TargetedBA}} & \parbox{0.9cm}{\tiny Trojan\\~\cite{Liu2018TrojaningAO}} & \parbox{1.0cm}{\tiny WaNet\\~\cite{Nguyen2021WaNetI}} & \parbox{0.9cm}{\tiny Dynamic\\~\cite{input-aware}} & \parbox{1.0cm}{\tiny Adap-\\Blend\\~\cite{revisitingbackdoor}} & \parbox{0.8cm}{\tiny Adap-\\Patch\\~\cite{revisitingbackdoor}} & \parbox{0.5cm}{ Clean} \\
\midrule
\multirow{2}{*}{ CIFAR-10} &  ACC & 0.936 & 0.934 & 0.939 & 0.926 & 0.941 & 0.933 & 0.936 & 0.937\\
&  ASR & 1.000 & 0.998 & 1.000 & 0.987 & 0.998 & 0.998 & 1.000 & -\\
\midrule
\multirow{2}{*}{ \shortstack{GTSRB}} &  ACC & 0.968 & 0.968 & 0.972 & 0.952 & 0.971 & 0.971 & 0.974 & 0.976\\
&  ASR & 1.000 & 0.996 & 1.000 & 0.986 & 1.000 & 0.995 & 0.982 & -\\
\bottomrule
\end{tabularx}
\end{adjustbox}
\label{tab:resnet18_acc_asr}
\vspace{-0.3cm}
\end{table}



%% file: tbls/mobilenet_acc_asr.tex
\begin{table}[ht]
\centering
\vspace{-0.1cm}
\captionof{table}{Accuracy and ASR on MobileNetV2.}
\begin{adjustbox}{max width=1\columnwidth}
\begin{tabularx}{\textwidth}{lXXXXXXXXX}
\toprule
\parbox{0.0cm}{} & \parbox{0.0cm}{} & \parbox{0.9cm}{ \tiny Badnets\\~\cite{Gu2017BadNetsIV}} & \parbox{0.9cm}{\tiny Blend\\~\cite{Chen2017TargetedBA}} & \parbox{0.9cm}{\tiny Trojan\\~\cite{Liu2018TrojaningAO}} & \parbox{1.0cm}{\tiny WaNet\\~\cite{Nguyen2021WaNetI}} & \parbox{0.9cm}{\tiny Dynamic\\~\cite{input-aware}} & \parbox{1.0cm}{\tiny Adap-\\Blend\\~\cite{revisitingbackdoor}} & \parbox{0.8cm}{\tiny Adap-\\Patch\\~\cite{revisitingbackdoor}} & \parbox{0.5cm}{ Clean} \\
\midrule
\multirow{2}{*}{ CIFAR-10} &  ACC & 0.905 & 0.906 & 0.901 & 0.907 & 0.905 & 0.898 & 0.902 & 0.906\\
&  ASR & 1.000 & 0.994 & 1.000 & 0.990 & 1.000 & 1.000 & 1.000 & -\\
\midrule
\multirow{2}{*}{ \shortstack{GTSRB}} &  ACC & 0.935 & 0.927 & 0.938 & 0.905 & 0.922 & 0.921 & 0.937 & 0.931\\
&  ASR & 1.000 & 0.994 & 1.000 & 0.991 & 1.000 & 1.000 & 1.000 & -\\
\bottomrule
\end{tabularx}
\end{adjustbox}
\label{tab:mobilenetv2_acc_asr}
\vspace{-0.3cm}
\end{table}

%% file: tbls/resnet_defenses_f1.tex
\begin{table}[ht]
\centering
\scriptsize
\captionof{table}{F1 scores of defense methods against backdoor attacks in CIFAR-10 and GTSRB. AVG stands for the average F1 score. }
\begin{adjustbox}{max width=1.2\columnwidth}
\begin{tabularx}{\textwidth}{lXXXXXXXXX}
\toprule
\parbox{0.0cm}{} & \parbox{0.0cm}{} & \parbox{0.9cm}{ \tiny Badnets\\~\cite{Gu2017BadNetsIV}} & \parbox{0.9cm}{\tiny Blend\\~\cite{Chen2017TargetedBA}} & \parbox{0.9cm}{\tiny Trojan\\~\cite{Liu2018TrojaningAO}} & \parbox{1.0cm}{\tiny WaNet\\~\cite{Nguyen2021WaNetI}} & \parbox{0.9cm}{\tiny Dynamic\\~\cite{input-aware}} & \parbox{1.0cm}{\tiny Adap-\\Blend\\~\cite{revisitingbackdoor}} & \parbox{0.8cm}{\tiny Adap-\\Patch\\~\cite{revisitingbackdoor}} & \parbox{0.5cm}{ AVG} \\
\midrule
\multirow{2}{*}{\tiny STRIP~\cite{Gao2019STRIPAD}} & \tiny cifar10 & \cellcolor{green!25}0.952 & \cellcolor{red!25}0.466 & \cellcolor{green!25}0.951 & \cellcolor{red!25}0.471 & \cellcolor{green!25}0.951 & \cellcolor{green!25}0.848 & \cellcolor{red!25}0.009 & \cellcolor{red!25}0.664\\
& \tiny gtsrb & \cellcolor{green!25}0.952 & \cellcolor{green!25}0.851 & \cellcolor{green!25}0.924 & \cellcolor{red!25}0.489 & \cellcolor{green!25}0.952 & \cellcolor{green!25}0.937 & \cellcolor{red!25}0.052 & \cellcolor{red!25}0.737\\
\midrule
\multirow{2}{*}{\tiny \shortstack{AC~\cite{Chen2018DetectingBA}}} & cifar10 & \cellcolor{green!25}1.000 & \cellcolor{green!25}0.946 & \cellcolor{green!25}1.000 & \cellcolor{green!25}0.883 & \cellcolor{green!25}0.978 & \cellcolor{green!25}1.000 & \cellcolor{red!25}0.000 & \cellcolor{green!25}0.830\\
& gtsrb & \cellcolor{red!25}0.000 & \cellcolor{red!25}0.000 & \cellcolor{red!25}0.000 & \cellcolor{red!25}0.000 & \cellcolor{red!25}0.000 & \cellcolor{red!25}0.000 & \cellcolor{red!25}0.000 & \cellcolor{red!25}0.000\\
\midrule
\multirow{2}{*}{\tiny Frequency~\cite{low-frequency}} & cifar10 & \cellcolor{green!25}1.000 & \cellcolor{green!25}0.921 & \cellcolor{green!25}1.000 & \cellcolor{red!25}0.141 & \cellcolor{green!25}0.981 & \cellcolor{green!25}0.921 & \cellcolor{red!25}0.784 & \cellcolor{green!25}0.821\\
& gtsrb & \cellcolor{green!25}0.854 & \cellcolor{green!25}0.812 & \cellcolor{green!25}0.854 & \cellcolor{red!25}0.361 & \cellcolor{red!25}0.792 & \cellcolor{green!25}0.814 & \cellcolor{red!25}0.679 & \cellcolor{red!25}0.738\\
\midrule
\multirow{2}{*}{\tiny SentiNet~\cite{Chou2018SentiNetDL}} & cifar10 & \cellcolor{green!25}0.952 & \cellcolor{red!25}0.114 & \cellcolor{red!25}0.291 & \cellcolor{red!25}0.170 & \cellcolor{red!25}0.596 & \cellcolor{red!25}0.121 & \cellcolor{green!25}0.957 & \cellcolor{red!25}0.457\\
& gtsrb & \cellcolor{green!25}0.952 & \cellcolor{red!25}0.434 & \cellcolor{green!25}0.952 & \cellcolor{red!25}0.484 & \cellcolor{red!25}0.721 & \cellcolor{red!25}0.792 & \cellcolor{green!25}0.975 & \cellcolor{red!25}0.759\\
\midrule
\multirow{2}{*}{\tiny CT~\cite{ct}} & cifar10 & \cellcolor{red!25}0.470 & \cellcolor{red!25}0.630 & \cellcolor{green!25}0.949 & \cellcolor{red!25}0.682 & \cellcolor{red!25}0.664 & \cellcolor{green!25}0.908 & \cellcolor{green!25}0.965 & \cellcolor{red!25}0.753\\
& gtsrb & \cellcolor{red!25}0.747 & \cellcolor{red!25}0.654 & \cellcolor{red!25}0.576 & \cellcolor{green!25}0.962 & \cellcolor{green!25}0.916 & \cellcolor{green!25}0.892 & \cellcolor{green!25}0.965 & \cellcolor{green!25}0.816\\
\midrule
\multirow{2}{*}{\tiny SS~\cite{Tran2018SpectralSI}} & cifar10 & \cellcolor{green!25}0.979 & \cellcolor{green!25}0.936 & \cellcolor{red!25}0.294 & \cellcolor{red!25}0.741 & \cellcolor{red!25}0.789 & \cellcolor{red!25}0.661 & \cellcolor{red!25}0.0208 & \cellcolor{red!25}0.632\\
& gtsrb & \cellcolor{green!25}0.829 & \cellcolor{green!25}0.807 & \cellcolor{green!25}0.965 & \cellcolor{red!25}0.530 & \cellcolor{green!25}0.875 & \cellcolor{red!25}0.538 & \cellcolor{red!25}0.681 & \cellcolor{red!25}0.746\\
\midrule
\multirow{2}{*}{\tiny SCAn~\cite{scan}} & cifar10 & \cellcolor{green!25}0.993 & \cellcolor{green!25}0.964 & \cellcolor{green!25}0.991 & \cellcolor{green!25}0.935 & \cellcolor{green!25}0.979 & \cellcolor{red!25}0.000 & \cellcolor{red!25}0.000 & \cellcolor{red!25}0.695\\
& gtsrb & \cellcolor{green!25}0.990 & \cellcolor{green!25}0.966 & \cellcolor{green!25}0.999 & \cellcolor{green!25}0.956 & \cellcolor{green!25}0.874 & \cellcolor{red!25}0.000 & \cellcolor{red!25}0.000 & \cellcolor{red!25}0.684\\
\midrule
\multirow{2}{*}{\tiny SPECTRE~\cite{SPECTRE}} & cifar10 & \cellcolor{green!25}0.990 & \cellcolor{green!25}0.990 & \cellcolor{green!25}0.991 & \cellcolor{green!25}0.839 & \cellcolor{green!25}0.991 & \cellcolor{green!25}0.938 & \cellcolor{green!25}0.865 & \cellcolor{green!25}0.943\\
& gtsrb & \cellcolor{green!25}0.957 & \cellcolor{green!25}0.954 & \cellcolor{green!25}0.968 & \cellcolor{red!25}0.000 & \cellcolor{green!25}0.976 & \cellcolor{red!25}0.000 & \cellcolor{red!25}0.000 & \cellcolor{red!25}0.551\\
\midrule
\multirow{2}{*}{\tiny \shortstack{\textsc{BProm} (10\%)}} & cifar10 & \cellcolor{green!25}1.000 & \cellcolor{green!25}1.000 & \cellcolor{green!25}1.000 & \cellcolor{green!25}1.000 & \cellcolor{green!25}1.000 & \cellcolor{green!25}1.000 & \cellcolor{green!25}1.000 & \cellcolor{green!25}\textbf{1.000}\\
& gtsrb & \cellcolor{green!25}1.000 & \cellcolor{green!25}1.000 & \cellcolor{green!25}1.000 & \cellcolor{green!25}1.000 & \cellcolor{green!25}1.000 & \cellcolor{green!25}1.000 & \cellcolor{green!25}1.000 & \cellcolor{green!25}\textbf{1.000}\\
\midrule
\multirow{2}{*}{\tiny \shortstack{\textsc{BProm} (5\%)}} & cifar10 & \cellcolor{green!25}1.000 & \cellcolor{green!25}1.000 & \cellcolor{green!25}1.000 & \cellcolor{green!25}1.000 & \cellcolor{green!25}1.000 & \cellcolor{green!25}1.000 & \cellcolor{green!25}1.000 & \cellcolor{green!25}1.000\\
& gtsrb & \cellcolor{green!25}1.000 & \cellcolor{green!25}0.965 & \cellcolor{green!25}1.000 & \cellcolor{green!25}1.000 & \cellcolor{green!25}1.000 & \cellcolor{green!25}1.000 & \cellcolor{green!25}1.000 & \cellcolor{green!25}0.995\\
\midrule
\multirow{2}{*}{\tiny \shortstack{\textsc{BProm} (1\%)}} & cifar10 & \cellcolor{green!25}1.000 & \cellcolor{green!25}1.000 & \cellcolor{green!25}1.000 & \cellcolor{green!25}1.000 & \cellcolor{green!25}1.000 & \cellcolor{green!25}1.000 & \cellcolor{green!25}1.000 & \cellcolor{green!25}1.000\\
& gtsrb & \cellcolor{green!25}1.000 & \cellcolor{red!25}0.782 & \cellcolor{green!25}1.000 & \cellcolor{green!25}1.000 & \cellcolor{green!25}1.000 & \cellcolor{green!25}1.000 & \cellcolor{green!25}1.000 & \cellcolor{green!25}0.969\\
\bottomrule
\end{tabularx}
\end{adjustbox}
\label{tab:defenses_f1}
\vspace{-0.3cm}
\end{table}

%% file: tbls/mobilenet_defenses_auroc.tex
\begin{table}[ht]
\centering
\scriptsize

\captionof{table}{AUROC of defenses on MobileNetV2 under backdoor attacks on CIFAR-10 and GTSRB. AVG stands for the average AUROC. }
\begin{adjustbox}{max width=1.2\columnwidth}
\begin{tabularx}{\textwidth}{lXXXXXXXXX}
\toprule
\parbox{0.0cm}{} & \parbox{0.0cm}{} & \parbox{0.9cm}{ \tiny Badnets\\~\cite{Gu2017BadNetsIV}} & \parbox{0.9cm}{\tiny Blend\\~\cite{Chen2017TargetedBA}} & \parbox{0.9cm}{\tiny Trojan\\~\cite{Liu2018TrojaningAO}} & \parbox{1.0cm}{\tiny WaNet\\~\cite{Nguyen2021WaNetI}} & \parbox{0.9cm}{\tiny Dynamic\\~\cite{input-aware}} & \parbox{1.0cm}{\tiny Adap-\\Blend\\~\cite{revisitingbackdoor}} & \parbox{0.8cm}{\tiny Adap-\\Patch\\~\cite{revisitingbackdoor}} & \parbox{0.5cm}{ AVG} \\

\midrule
\multirow{2}{*}{\tiny STRIP~\cite{Gao2019STRIPAD}} & cifar10 & \cellcolor{red!25}0.739 & \cellcolor{green!25}0.833 & \cellcolor{green!25}0.957 & \cellcolor{red!25}0.473 & \cellcolor{green!25}0.987 & \cellcolor{green!25}0.987 & \cellcolor{green!25}0.952 & \cellcolor{green!25}0.847\\
& gtsrb & \cellcolor{red!25}0.798 & \cellcolor{green!25}0.873 & \cellcolor{red!25}0.745 & \cellcolor{red!25}0.489 & \cellcolor{green!25}0.998 & \cellcolor{green!25}0.981 & \cellcolor{green!25}0.999 & \cellcolor{green!25}0.840\\
\midrule
\multirow{2}{*}{\shortstack{\tiny AC~\cite{Chen2018DetectingBA}}} & cifar10 & \cellcolor{red!25}0.364 & \cellcolor{red!25}0.398 & \cellcolor{green!25}0.996 & \cellcolor{green!25}0.889 & \cellcolor{red!25}0.425 & \cellcolor{red!25}0.390 & \cellcolor{green!25}1.000 & \cellcolor{red!25}0.637\\
& gtsrb & \cellcolor{red!25}0.225 & \cellcolor{red!25}0.309 & \cellcolor{red!25}0.263 & \cellcolor{red!25}0.355 & \cellcolor{red!25}0.288 & \cellcolor{red!25}0.576 & \cellcolor{red!25}0.263 & \cellcolor{red!25}0.326\\
\midrule
\multirow{2}{*}{\tiny Frequency~\cite{low-frequency}} & cifar10 & \cellcolor{green!25}1.000 & \cellcolor{green!25}0.996 & \cellcolor{green!25}1.000 & \cellcolor{green!25}0.999 & \cellcolor{green!25}0.970 & \cellcolor{green!25}0.996 & \cellcolor{green!25}1.000 & \cellcolor{green!25}0.994\\
& gtsrb & \cellcolor{green!25}1.000 & \cellcolor{green!25}0.973 & \cellcolor{green!25}1.000 & \cellcolor{red!25}0.777 & \cellcolor{green!25}0.960 & \cellcolor{green!25}0.973 & \cellcolor{green!25}1.000 & \cellcolor{green!25}0.955\\
\midrule
\multirow{2}{*}{\tiny CT~\cite{ct}} & cifar10 & \cellcolor{green!25}0.999 & \cellcolor{green!25}0.996 & \cellcolor{green!25}0.999 & \cellcolor{green!25}0.943 & \cellcolor{green!25}0.985 & \cellcolor{green!25}0.992 & \cellcolor{green!25}0.802 & \cellcolor{green!25}0.959\\
& gtsrb & \cellcolor{green!25}0.993 & \cellcolor{green!25}0.984 & \cellcolor{green!25}0.998 & \cellcolor{green!25}0.852 & \cellcolor{green!25}0.985 & \cellcolor{green!25}0.985 & \cellcolor{red!25}0.616 & \cellcolor{green!25}0.916\\
\midrule
\multirow{2}{*}{\tiny SS~\cite{Tran2018SpectralSI}} & cifar10 & \cellcolor{red!25}0.439 & \cellcolor{red!25}0.428 & \cellcolor{red!25}0.375 & \cellcolor{red!25}0.381 & \cellcolor{red!25}0.426 & \cellcolor{red!25}0.377 & \cellcolor{red!25}0.442 & \cellcolor{red!25}0.410\\
& gtsrb & \cellcolor{red!25}0.492 & \cellcolor{red!25}0.492 & \cellcolor{red!25}0.492 & \cellcolor{red!25}0.492 & \cellcolor{red!25}0.487 & \cellcolor{red!25}0.492 & \cellcolor{red!25}0.487 & \cellcolor{red!25}0.491\\
\midrule
\multirow{2}{*}{\tiny SCAn~\cite{scan}} & cifar10 & \cellcolor{green!25}0.991 & \cellcolor{green!25}0.921 & \cellcolor{green!25}0.953 & \cellcolor{green!25}0.926 & \cellcolor{green!25}0.988 & \cellcolor{green!25}0.981 & \cellcolor{green!25}0.926 & \cellcolor{green!25}0.955\\
& gtsrb & \cellcolor{green!25}0.999 & \cellcolor{green!25}0.979 & \cellcolor{green!25}0.969 & \cellcolor{green!25}0.952 & \cellcolor{green!25}0.986 & \cellcolor{green!25}0.969 & \cellcolor{green!25}0.976 & \cellcolor{green!25}0.976\\
\midrule
\multirow{2}{*}{\tiny SPECTRE~\cite{SPECTRE}} & cifar10 & \cellcolor{green!25}0.857 & \cellcolor{red!25}0.376 & \cellcolor{green!25}0.876 & \cellcolor{red!25}0.534 & \cellcolor{green!25}0.897 & \cellcolor{red!25}0.510 & \cellcolor{red!25}0.376 & \cellcolor{red!25}0.632\\
& gtsrb & \cellcolor{green!25}0.911 & \cellcolor{red!25}0.699 & \cellcolor{red!25}0.797 & \cellcolor{red!25}0.597 & \cellcolor{red!25}0.595 & \cellcolor{red!25}0.617 & \cellcolor{red!25}0.581 & \cellcolor{red!25}0.685\\
\midrule
\multirow{2}{*}{\tiny \shortstack{\textsc{BProm} (10\%)}} & cifar10 & \cellcolor{green!25}1.000 & \cellcolor{green!25}1.000 & \cellcolor{green!25}1.000 & \cellcolor{green!25}1.000 & \cellcolor{green!25}1.000 & \cellcolor{green!25}1.000 & \cellcolor{green!25}1.000 & \cellcolor{green!25}\textbf{1.000}\\
& gtsrb & \cellcolor{green!25}1.000 & \cellcolor{green!25}0.999 & \cellcolor{green!25}1.000 & \cellcolor{green!25}1.000 & \cellcolor{green!25}1.000 & \cellcolor{green!25}1.000 & \cellcolor{green!25}1.000 & \cellcolor{green!25}\textbf{1.000}\\
\bottomrule
\end{tabularx}
\end{adjustbox}
\label{tab:mobilenet_defenses_auroc}
\end{table}

%% file: tbls/mobilenet_defenses_f1.tex
\begin{table}[ht]
\centering
\scriptsize

\captionof{table}{F1 score of defenses on MobileNetV2 under backdoor attacks on CIFAR-10 and GTSRB. AVG stands for the average F1 score. }
\begin{adjustbox}{max width=1.2\columnwidth}
\begin{tabularx}{\textwidth}{lXXXXXXXXX}
\toprule
\parbox{0.0cm}{} & \parbox{0.0cm}{} & \parbox{0.9cm}{ \tiny Badnets\\~\cite{Gu2017BadNetsIV}} & \parbox{0.9cm}{\tiny Blend\\~\cite{Chen2017TargetedBA}} & \parbox{0.9cm}{\tiny Trojan\\~\cite{Liu2018TrojaningAO}} & \parbox{1.0cm}{\tiny WaNet\\~\cite{Nguyen2021WaNetI}} & \parbox{0.9cm}{\tiny Dynamic\\~\cite{input-aware}} & \parbox{1.0cm}{\tiny Adap-\\Blend\\~\cite{revisitingbackdoor}} & \parbox{0.8cm}{\tiny Adap-\\Patch\\~\cite{revisitingbackdoor}} & \parbox{0.5cm}{\tiny AVG} \\
\midrule
\multirow{2}{*}{\tiny STRIP~\cite{Gao2019STRIPAD}} & cifar10 & \cellcolor{red!25}0.552 & \cellcolor{red!25}0.673 & \cellcolor{green!25}0.916 & \cellcolor{red!25}0.122 & \cellcolor{green!25}0.939 & \cellcolor{green!25}0.943 & \cellcolor{green!25}0.999 & \cellcolor{red!25}0.735\\
& gtsrb & \cellcolor{red!25}0.513 & \cellcolor{green!25}0.800 & \cellcolor{red!25}0.442 & \cellcolor{red!25}0.148 & \cellcolor{green!25}0.954 & \cellcolor{green!25}0.937 & \cellcolor{green!25}0.955 & \cellcolor{red!25}0.678\\
\midrule
\multirow{2}{*}{\tiny \shortstack{AC~\cite{Chen2018DetectingBA}}} & cifar10 & \cellcolor{red!25}0.000 & \cellcolor{red!25}0.000 & \cellcolor{green!25}0.998 & \cellcolor{green!25}0.942 & \cellcolor{red!25}0.000 & \cellcolor{red!25}0.000 & \cellcolor{green!25}1.000 & \cellcolor{red!25}0.420\\
& gtsrb & \cellcolor{red!25}0.000 & \cellcolor{red!25}0.000 & \cellcolor{red!25}0.000 & \cellcolor{red!25}0.000 & \cellcolor{red!25}0.000 & \cellcolor{red!25}0.000 & \cellcolor{red!25}0.000 & \cellcolor{red!25}0.000\\
\midrule
\multirow{2}{*}{\tiny Frequency~\cite{low-frequency}} & cifar10 & \cellcolor{green!25}0.983 & \cellcolor{green!25}0.922 & \cellcolor{green!25}0.984 & \cellcolor{green!25}0.981 & \cellcolor{red!25}0.139 & \cellcolor{green!25}0.92 & \cellcolor{green!25}0.983 & \cellcolor{green!25}0.844\\
& gtsrb & \cellcolor{green!25}0.865 & \cellcolor{green!25}0.825 & \cellcolor{green!25}0.864 & \cellcolor{red!25}0.392 & \cellcolor{green!25}0.804 & \cellcolor{green!25}0.826 & \cellcolor{green!25}0.865 & \cellcolor{red!25}0.777\\
\midrule
\multirow{2}{*}{\tiny CT~\cite{ct}} & cifar10 & \cellcolor{green!25}0.998 & \cellcolor{green!25}0.988 & \cellcolor{green!25}0.999 & \cellcolor{green!25}0.936 & \cellcolor{green!25}0.981 & \cellcolor{green!25}0.992 & \cellcolor{red!25}0.753 & \cellcolor{green!25}0.950\\
& gtsrb & \cellcolor{green!25}0.976 & \cellcolor{green!25}0.967 & \cellcolor{green!25}0.992 & \cellcolor{green!25}0.810 & \cellcolor{green!25}0.960 & \cellcolor{green!25}0.968 & \cellcolor{red!25}0.378 & \cellcolor{green!25}0.864\\
\midrule
\multirow{2}{*}{\tiny SS~\cite{Tran2018SpectralSI}} & cifar10 & \cellcolor{red!25}0.230 & \cellcolor{red!25}0.222 & \cellcolor{red!25}0.141 & \cellcolor{red!25}0.176 & \cellcolor{red!25}0.218 & \cellcolor{red!25}0.173 & \cellcolor{red!25}0.232 & \cellcolor{red!25}0.199\\
& gtsrb & \cellcolor{red!25}0.278 & \cellcolor{red!25}0.279 & \cellcolor{red!25}0.278 & \cellcolor{red!25}0.278 & \cellcolor{red!25}0.272 & \cellcolor{red!25}0.278 & \cellcolor{red!25}0.278 & \cellcolor{red!25}0.277\\
\midrule
\multirow{2}{*}{\tiny SCAn~\cite{scan}} & cifar10 & \cellcolor{green!25}0.938 & \cellcolor{green!25}0.908 & \cellcolor{green!25}0.997 & \cellcolor{green!25}0.920 & \cellcolor{green!25}0.987 & \cellcolor{green!25}0.981 & \cellcolor{green!25}0.974 & \cellcolor{green!25}0.958\\
& gtsrb & \cellcolor{red!25}0.642 & \cellcolor{red!25}0.769 & \cellcolor{green!25}1.000 & \cellcolor{green!25}0.818 & \cellcolor{green!25}0.968 & \cellcolor{red!25}0.450 & \cellcolor{red!25}0.335 & \cellcolor{red!25}0.712\\
\midrule
\multirow{2}{*}{\tiny SPECTRE~\cite{SPECTRE}} & cifar10 & \cellcolor{red!25}0.774 & \cellcolor{red!25}0.676 & \cellcolor{red!25}0.125 & \cellcolor{red!25}0.282 & \cellcolor{red!25}0.775 & \cellcolor{red!25}0.252 & \cellcolor{red!25}0.169 & \cellcolor{red!25}0.436\\
& gtsrb & \cellcolor{green!25}0.897 & \cellcolor{red!25}0.401 & \cellcolor{red!25}0.748 & \cellcolor{red!25}0.358 & \cellcolor{red!25}0.356 & \cellcolor{red!25}0.388 & \cellcolor{red!25}0.338 & \cellcolor{red!25}0.498\\
\midrule
\multirow{2}{*}{\tiny \shortstack{\textsc{BProm} (10\%)}} & cifar10 & \cellcolor{green!25}1.000 & \cellcolor{green!25}1.000 & \cellcolor{green!25}1.000 & \cellcolor{green!25}0.967 & \cellcolor{green!25}1.000 & \cellcolor{green!25}1.000 & \cellcolor{green!25}1.000 & \cellcolor{green!25}\textbf{0.995}\\
& gtsrb & \cellcolor{green!25}1.000 & \cellcolor{green!25}0.965 & \cellcolor{green!25}1.000 & \cellcolor{green!25}1.000 & \cellcolor{green!25}0.965 & \cellcolor{green!25}0.965 & \cellcolor{green!25}1.000 & \cellcolor{green!25}\textbf{0.985}\\
\bottomrule
\end{tabularx}
\end{adjustbox}
\label{tab:mobilenet_defenses_f1}
\end{table}

%% file: tbls/imagenet_auc.tex
\begin{table}[ht]
\centering
\caption{AUROC of \textsc{BProm} and other defense methods against various backdoor attacks on ImageNet. AVG stands for the average AUROC.}
\label{tab:imagenet_auc}
\begin{tabular}{cccccc}
\toprule
  & Badnets & Trojan & Adap-Blend & Adap-Patch & AVG \\
\cmidrule{1-6}
CD~\citep{cd}        & 0.7954 & 0.9424 & 0.6648 & 0.5842 & 0.7467 \\
\cmidrule{1-6}
SCALE-UP~\citep{scaleup} & 0.9912 & 0.6556 & 0.3971 & 0.3339 & 0.5944 \\
\cmidrule{1-6}
STRIP~\citep{Gao2019STRIPAD}   & 0.0500 & 0.0500 & 0.5244 & 0.5500 & 0.2936 \\
\cmidrule{1-6}
\textsc{BProm} (10\%) & 1.0000 & 1.0000 & 0.9986 & 0.8296 & 0.9570 \\
\bottomrule
\end{tabular}
\end{table}